\documentclass[10pt]{article} 
\usepackage[preprint]{tmlr}
\usepackage[utf8]{inputenc}
\usepackage{amsfonts}
\usepackage{amsmath}
\usepackage{amssymb}
\usepackage{amsthm}
\usepackage{mathtools}
\usepackage{rotating}
\usepackage{enumitem}
\usepackage{todonotes}
\usepackage[hidelinks]{hyperref}
\usepackage{cleveref}
\crefname{subsection}{Subsection}{Subsections}

\usepackage{longtable}
\usepackage{pgfplotstable}
\usepackage{booktabs}
\usepackage{float}
\usepackage{subcaption}
\usepackage{datatool}
\usepackage{orcidlink}
\usepackage{acro}

\DeclareAcronym{ML}{
  short=ML,
  long=machine learning,
}
\DeclareAcronym{ID}{
  short=ID,
  long=intrinsic dimension,
}
\DeclareAcronym{NID}{
  short=NID,
  long=normalized intrinsic dimensionality,
}
\DeclareAcronym{SGD}{
  short=SGD,
  long=stochastic gradient decent,
}
\DeclareAcronym{COD}{
  short=COD,
  long=curse of dimensionality,
}
\newcommand{\cod}{\emph{\acl{COD}}}

\newcommand{\eg}{\textit{e}.\textit{g}.\ }

\newtheorem{definition}{Definition}

\usepackage{usebib}
\bibinput{bibliography2}
\newcommand{\citetitle}[1]{\usebibentry{#1}{title}}

\let\oldtodo\todo
\renewcommand{\todo}[1]{\oldtodo[inline]{#1}}

\newcommand{\longtextorientation}{270}
\newif\ifexplanation
\explanationtrue
\newcommand{\ddash}{\textemdash}

\newcommand{\showexplanation}[1]{\ifexplanation\ddash #1\else\!\!\fi}

\newcommand{\rotatelongtext}[1]{
  \rotatebox{\longtextorientation}{#1}
}

\newcommand{\datasetcategory}{
  \multicolumn{11}{c|}{data set}
}
\newcommand{\datasetsubcategories}{
  \multicolumn{7}{c|}{\rotatelongtext{availability}} &
  \multicolumn{4}{c|}{\rotatelongtext{transformation }}
}
\newcommand{\datasetsubsubcategories}{
  \multicolumn{2}{c|}{\rotatelongtext{metadata}} &
  \multicolumn{5}{c|}{\rotatelongtext{download}} &
  \multicolumn{2}{c|}{\rotatelongtext{preprocessing }} &
  \multicolumn{2}{c|}{\rotatelongtext{selection}}
}
\newcommand{\datasetsubsubsubcategories}{
  \rotatelongtext{D1\showexplanation{format not documented}} &
  \rotatelongtext{D2\showexplanation{version not specified}} &
  \rotatelongtext{D3\showexplanation{no direct access}} &
  \rotatelongtext{D4\showexplanation{access not possible}} &
  \rotatelongtext{D5\showexplanation{privacy restricted}} &
  \rotatelongtext{D6\showexplanation{license restricted}} &
  \rotatelongtext{D7\showexplanation{on request only}} &
  \rotatelongtext{D8\showexplanation{manual steps}} &
  \rotatelongtext{D9\showexplanation{incomplete description}} &
  \rotatelongtext{D10\showexplanation{train/test splits unclear}} &
  \rotatelongtext{D11\showexplanation{number of samples not documented}}
}

\newcommand{\softwarecategory}{
  \multicolumn{19}{c|}{software}
}
\newcommand{\softwaresubcategories}{
  \multicolumn{5}{c|}{\rotatelongtext{environment }} &
  \multicolumn{5}{c|}{\rotatelongtext{usage}} &
  \multicolumn{9}{c|}{\rotatelongtext{source code}}
}
\newcommand{\softwaresubsubcategories}{
  \multicolumn{3}{c|}{\rotatelongtext{dependencies}} &
  \multicolumn{2}{c|}{\rotatelongtext{variables}} &
  \multicolumn{3}{c|}{\rotatelongtext{documentation }} &
  \multicolumn{2}{c|}{\rotatelongtext{scripts}} &
  \multicolumn{5}{c|}{\rotatelongtext{bugs}} &
  \multicolumn{4}{c|}{\rotatelongtext{experiments}}
}
\newcommand{\softwaresubsubsubcategories}{
  \rotatelongtext{S1\showexplanation{exact version not documented}} &
  \rotatelongtext{S2\showexplanation{specified version not available anymore}} &
  \rotatelongtext{S3\showexplanation{necessary hardware unavailable}} &
  \rotatelongtext{S4\showexplanation{seeds not set}} &
  \rotatelongtext{S5\showexplanation{important values unclear}} &
  \rotatelongtext{S6\showexplanation{not up to date}} &
  \rotatelongtext{S7\showexplanation{necessary arguments not clear}} &
  \rotatelongtext{S8\showexplanation{missing hyperparameters}} &
  \rotatelongtext{S9\showexplanation{incomplete train/test scripts}} &
  \rotatelongtext{S10\showexplanation{unclear which version was used}} &
  \rotatelongtext{S11\showexplanation{never fixed}} &
  \rotatelongtext{S12\showexplanation{issue solutions not applied}} &
  \rotatelongtext{S13\showexplanation{fix distributed through other channels}} &
  \rotatelongtext{S14\showexplanation{api changes}} &
  \rotatelongtext{S15\showexplanation{out of memory}} &
  \rotatelongtext{S16\showexplanation{one missing}} &
  \rotatelongtext{S17\showexplanation{all missing}} &
  \rotatelongtext{S18\showexplanation{hyperparameter search not included}} &
  \rotatelongtext{S19\showexplanation{only general idea implemented}}
}

\newcommand{\resultscategory}{
  \multicolumn{6}{c|}{computational result}
}
\newcommand{\resultssubcategories}{
  \rotatelongtext{model} &
  \multicolumn{5}{c|}{\rotatelongtext{predictions}}
}
\newcommand{\resultssubsubcategories}{
  \rotatelongtext{ } &
  \multicolumn{5}{c|}{ }
}
\newcommand{\resultssubsubsubcategories}{
  \rotatelongtext{R1\showexplanation{no model weights}} &
  \rotatelongtext{R2\showexplanation{small deviations}} &
  \rotatelongtext{R3\showexplanation{strong differences in parts}} &
  \rotatelongtext{R4\showexplanation{strong differences everywhere}} &
  \rotatelongtext{R5\showexplanation{weak statistics}} &
  \rotatelongtext{R6\showexplanation{no predictions}}
}

\pgfplotstableread[col sep = &, header = false]{tables/context.dat}\contextdata

\newcommand{\tablebuilder}[3]{
  \pgfplotstabletypeset[
  every head row/.style={
    output empty row,
    before row={
      {#1}
    }
  },
  columns={#2},
  string type,
  column type/.add={|}{},
  every last column/.style={column type/.add={}{|}},
  header=false,
  every last row/.style={
    after row=\bottomrule
  },
  display columns/0/.style={string type}
  ]\contextdata
}

\newcommand{\datasettable}[1]{
  \begin{center}
    #1
    \tablebuilder{
      \toprule &
      \datasetcategory \\
      \midrule &
      \datasetsubcategories \\
      \midrule &
      \datasetsubsubcategories \\
      \midrule &
      \datasetsubsubsubcategories \\
      \midrule
    }{
      {0,1,2,3,4,5,6,7,8,9,10,11}
    }{
      dataset.tex
    }
  \end{center}
}

\newcommand{\softwaretable}[1]{
  \begin{center}
    #1
    \tablebuilder{
      \toprule &
      \softwarecategory \\
      \midrule &
      \softwaresubcategories \\
      \midrule &
      \softwaresubsubcategories \\
      \midrule &
      \softwaresubsubsubcategories \\
      \midrule
    }{
      {0,12,13,14,15,16,17,18,19,20,21,22,23,24,25,26,27,28,29,30}
    }{
      software.tex
    }
  \end{center}
}

\newcommand{\resultstable}[1]{
  \begin{center}
    #1
    \tablebuilder{
      \toprule &
      \resultscategory \\
      \midrule &
      \resultssubcategories \\
      \midrule &
      \resultssubsubcategories \\
      \midrule &
      \resultssubsubsubcategories \\
      \midrule
    }{
      {0,31,32,33,34,35,36}
    }{
      results.tex
    }
  \end{center}
}

\newcommand{\contexttable}{
\tablebuilder{
  \toprule &
  \datasetcategory &
  \softwarecategory &
  \resultscategory \\
  \midrule &
  \datasetsubcategories &
  \softwaresubcategories &
  \resultssubcategories \\
  \midrule &
  \datasetsubsubcategories &
  \softwaresubsubcategories &
  \resultssubsubcategories \\
  \midrule &
  \datasetsubsubsubcategories &
  \softwaresubsubsubcategories &
  \resultssubsubsubcategories \\
  \midrule
}{
  {0,1,2,3,4,5,6,7,8,9,10,11,12,13,14,15,16,17,18,19,20,21,22,23,24,25,26,27,28,29,30,31,32,33,34,35,36}
}{
  context.tex
}
}

\DTLloaddb[keys={data_set_name,nodes,edges,features,paper_names}]{data_set_sizes_grouped}{%
  tables/data_set_sizes_grouped.csv
}
\let\cref\Cref

\begin{document}
\def\month{MM}
\def\year{YYYY}
\def\openreview{\url{https://openreview.net/forum?id=XXXX}}
%
%
\title{Reproducibility and Geometric Intrinsic Dimensionality:\\An Investigation on Graph
  Neural Network Research}
\newcommand{\kde}{Knowledge \& Data Engineering Group,
  University of Kassel, Kassel,  Germany}
\newcommand{\iteg}{Interdisciplinary Research Center for Information System
  Design, University of Kassel, Kassel, Germany}
\newcommand{\ismll}{Information Systems and Machine Learning Lab,
  University of Hildesheim, Hildesheim, Germany}
\newcommand{\isi}{Institute of Computer Science,
  University of Hildesheim, Hildesheim, Germany}
\author{\name Tobias Hille\orcidlink{0000-0001-7813-9799}
  \email hille@cs.uni-kassel.de\\
  \addr \kde\\
  \iteg
  \AND
  \name Maximilian Stubbemann\orcidlink{0000-0003-1579-1151}
  \email stubbemann@ismll.de\\
  \addr \ismll \\ \kde\\ \iteg
  \AND
  \name Tom Hanika\orcidlink{0000-0002-4918-6374}
  \email tom.hanika@uni-hildesheim.de\\
  \addr \isi\\
  \kde
}
%
\setcounter{tocdepth}{1}

\maketitle
\begin{abstract}
  Difficulties in replication and reproducibility of empirical evidences in
  machine learning research have become a prominent topic in recent years.
  Ensuring that machine learning research results are sound and reliable
  requires reproducibility, which verifies the reliability of research findings
  using the same code and data.
  This promotes open and accessible research, robust experimental workflows, and
  the rapid integration of new findings.
  Evaluating the degree to which research publications support these different
  aspects of reproducibility is one goal of the present work.
  For this we introduce an ontology of reproducibility in machine learning and
  apply it to methods for graph neural networks.\\
  Building on these efforts we turn towards another critical challenge in machine
  learning, namely the \cod, which poses challenges in
  data collection, representation, and analysis, making it harder to find
  representative data and impeding the training and inference processes.
  Using the closely linked concept of geometric intrinsic dimension we
  investigate to which extend the used machine learning models are influenced by
  the intrinsic dimension of the data sets they are trained on.
\end{abstract}


\section{Introduction}
\Ac{ML} is a rapidly evolving field that has made significant contributions to
numerous industries.
In view of its considerable impact, it also becomes apparent how difficult it
isto replicate and reproduce empirical findings in the field of \ac{ML}.
Therefore, reproducibility in \ac{ML} has become an important topic in its own
right in recent years.
Reproducibility, defined as the ability of a researcher to duplicate the results
of a prior study using the same materials as the original investigator, is
critical to ensuring the validity and reliability of research findings.
It promotes transparency, allows for verification of results, and fosters
confidence in the scientific community.
Despite its importance, achieving reproducibility in \ac{ML} research is
challenging due to several barriers.
One of the main difficulties is the implementation of the exact experimental and
computational procedures as described in the original work.
The resulting layers of complexity become particularly apparent when the used
computational frameworks continually update and rise and fall in popularity and
levels of maintenance.
Another major challenge is the inherent instability of results.
This is influenced by a multitude of factors such as the amount of data
available, the computational resources at hand, the determination of
hyperparameters, and the inherent randomness of the training process.
In this context, it is even more difficult to assess the influence of
uncontrolled epistemic uncertainties, such as the intrinsic dimensionality.
Several guidelines, originating from conferences, workshops, and coding
frameworks, provide recommendations and tools that help researchers and authors
in this regard~\citep{ reproducibility_checklist, reproducibility_workshop_iclr,
  pytorch_lightning}.
However, these are often not very detailed or allow limited structural
evaluation and comparability of reproducibility.
Moreover, as several authors ascertain a lack of standard terminology for
reproducibility (within \ac{ML}) which hinders the emergence of an unified
evaluation framework~\citep{Tatman2018APT,Bouthillier2019UnreproducibleRI}.

This paper proposes a comprehensive and in-depth framework for the study of
reproducibility in the research area of graph neural networks.
The challenges associated with a data set, a method and a result are analysed in
terms of their significance for computational reproducibility.
A multi-stage selection process identified six scientific papers for which we
studied and adapted our framework.
With their help, we explore and demonstrate the limits and difficulties of
reproducibility.
This results in a new ontology for scientific reproducibility that generalizes
to the realm of machine learning as a whole.

A second major challenge for the reproducibility of high-dimensional \ac{ML}
results is the occurrence of epistemic uncertainties.
This is particularly the case when an attempt is made to transfer a result to
new data or use cases.
A particular instance of this uncertainty is the umbrella term \cod.
This is based on various mathematical observations in high-dimensional spaces
that are generally not addressed by \ac{ML} studies.
A geometric approach towards understanding the \cod~was established by
V.~Pestov.
He proved that the concentration of measure
phenomenon~\cite{Milman1988TheHO,Milman2000TopicsIA} contributes to the overall
\cod~\citep{Pestov1999OnTG, Pestov2007IntrinsicDO, Pestov2007AnAA,
  Pestov2010IntrinsicD, Pestov2010IndexabilityCA}.
His approach was adapted towards a practical computable function for estimating
the \ac{ID} of a geometric data set~\citep{Hanika2022IntrinsicDO}.
This result was further improved with regard to its applicability to large data
sets~\citep{ Stubbemann2022IntrinsicDF, Stubbemann2023SelectingFB}.

With regard to reproducibility, we investigate the influence of the \ac{ID} on
the \ac{ML} training process.
In particular, we experiment with \ac{ID}-based feature selection, as it
provides a straightforward method to manipulate the \ac{ID} of a data set.
As we hypothesize that training methods are susceptible to \ac{ID}-changes in
the underlying training data set, we apply different \ac{ML} methods to the same
manipulated data sets.
We thereby study the impact of altering the intrinsic dimension of graph data
sets for all six reproduced graph neural network methods.

Although there are studies on these theoretical and practical aspects, the
present work aims to bridge the gap between them by focusing on reproducibility
and the intrinsic dimension within a geometric understanding.
Altogether, our work contributes to improving the quality and reliability of
\ac{ML} research, ultimately benefiting the broader scientific community and
industry applications.

To summarize our contributions:
\begin{itemize}
\item \textbf{We introduce an ontology of reproducibility in Machine
    Learning}~(\cref{sec:ontology}).
\item \textbf{We consider about 100 publications from the field of graph
    neural networks and reproduce six of them
    extensively}~(\cref{sec:survey}).
\item \textbf{We investigate how the change of the (geometric)
    intrinsic dimension in data sets effects the performance of the
    six reproduced methods}~(\cref{sec:dimensionality}).
\end{itemize}


\section{Related Work}
\label{sec:related_work}
\subsection*{Reproducibility and Replicability}
\label{sec:related_work:reproducibility_and_replicability}
Several publications have investigated the general
state~\citep{repr_and_repl_in_science} and
challenges~\citep{irreproducibility_special_nature} of reproducibility and
replicability in science.
There are also works that looked more specific into these questions in the
field of computer science~\citep{computational_reproducibility} and its
sub-field of machine learning~\citep{Raff2019AST, Liu2020OnTR,
  Chen2022TowardsTR}.
In recent years a growing number of conferences include dedicated tracks for
reproducibility efforts or specific workshops~\citep{
  Stodden2013SettingTD,
  reproducibility_workshop_iclr}.
The knowledge collected there is now available in general
reports~\citep{improving_reproducibility} and straightforward
checklists~\citep{reproducibility_checklist}.
Related to this more and more journals and publisher provide specific editorial
policies~\citep{Casadevall2010ReproducibleS, editorial_policy_nature} to help
authors in that regard.
Efforts for reproducing and replicating past works from broad range of research
fields concentrate in some dedicated journals~\citep{rescience}, in which
publications from further back are also of interest.%
\footnote{See the \textit{ten years challenge}
  \url{http://rescience.github.io/ten-years/}}
Beyond the space of academic publication there are of course similar efforts
made by the programming community~\citep{research_code_publishing,
  practice_for_reproducibility}.
Specifically machine learning engineering teams and individuals build
frameworks~\citep{pytorch_lightning} and templates
\citep{lightning_hydra_template} for streamlining the process of setting up a
reproducible machine learning experiment.
Recent investigations showed, that the choice of the used machine learning
framework and its version~\citep{
  Pham2020ProblemsAO,
  Shahriari2022HowDD}
or commercially available platforms providing related
services~\citep{Gundersen2022DoML} can have a significant impact on the
reproducibility properties of the research code.
There are, motivated by practical concerns, surveys that
investigate directly the availability and operability of research
code~\citep{Warren2015RepeatabilityAB}.
Few works additionally try to construct a taxonomy of those
reproducibility properties~\citep{
  Goodman2016WhatDR,
  Marwick2017AssessingR,
  Tatman2018APT,
  Bouthillier2019UnreproducibleRI}.
The accompanying discussions often emphasize the confusing terminology~\citep{
  Peng2011ReproducibleRI,
  Plesser2018ReproducibilityVR,
  Gundersen2020TheFP}.
The provided taxonomies usually consist of a shallow hierarchy of different
levels of reproducibility which are characterized by high-level features of the
submissions that have to be assessed.
In most cases the process of evaluating is guided by only a few questions.
As such they give researchers and reviewers not that much guidance when
evaluating the degree of reproducibility.
However there are publications that go more into detail when analysing factors
and variables that influence reproducibility~\citep{
  Ivie2018ReproducibilityIS,
  Gundersen2018OnRA,
  Gundersen2018StateOT,
  Gundersen2022SourcesOI}.
This focus on central aspects of computational reproduciblity can
also be found in the present work.

\subsection*{Intrinsic Dimension and Feature Selection}
\label{sec:related_work:intrinsic_dimension_and_feature_selection}
The term \textit{\acf{ID}} has multiple slightly different
meanings in related sub-fields of machine learning.
They share the motivating aspect of using the value of the \ac{ID}
of data as a proxy for gaining evidence on how the data is structured.
One prominent usage of the term is for specifying the often approximated
dimension of a hypothetical embedded manifold in the data space which describes
almost all samples with sufficient accuracy~\citep{
  Hein2005IntrinsicDE,
  Tatti2006WhatIT}.
This notion of \ac{ID} can be used to motivate a variety of estimators, for
example based on sampling around data point
neighborhoods~\citep{Kim2016MinimaxRF}.
Those estimators give rise to different feature selection methods~\citep{
  Traina2010FastFS,
  Mo2012FractalBasedID,
  Suryakumar2013InfluenceOM,
  Golay2016FeatureSF}, occasionally based on gradients to learn an embedding
with the desired properties~\citep{Pope2021TheID}.
However these algorithms do not help to decide if and to what extent the data
set is affected by the \cod~and the related
concentration phenomena~\citep{Franois2007TheCO, Houle2013DimensionalityDD}.

In contrast, the \acl{ID} for data by Pestov~\citep{
  Pestov1999OnTG,
  Pestov2007IntrinsicDO,
  Pestov2010IntrinsicD,
  Pestov2011IsTK} gives an axiomatic approach on quantifying the influence of
the \cod~directly.
It links the latter to the phenomenon of concentration of measure and builds on
the axiomatization of the latter by Gromov and Milman~\citep{
  Gromov1983ATA,
  Milman1988TheHO,
  Milman2000TopicsIA}.
In this construction it amounts to computing the set of all real-valued
1-Lipschitz functions on a given metric space as potential features.
These mathematical works give rise to the intuitive view on the \cod~as the
phenomenon of features concentrating near their means or medians, so that
algorithms are therefore not able no discriminate the data.
However this approach by itself was computationally infeasible until the
introduction of the \acl{ID} of \textit{geometric data sets}~\citep{
  Hanika2022IntrinsicDO}.
Here a set of features is defined beforehand.
Later publications provided algorithms for computation or approximation of this
\ac{ID}~\citep{Stubbemann2022IntrinsicDF} and its application to feature
selection~\citep{Stubbemann2023SelectingFB}, even for large-scale data sets.

\subsection*{Studies Regarding Influence of Data on Model Behavior}
\label{sec:related_work:study_influence_data_model_behavior}
Machine learning models are heavily influenced by the quality and nature of the
input data they are trained on. This relationship has been extensively studied
in various contexts, leading to significant advancements and challenges in the
field.
A large body of literature is concerned with in influence of \textit{simple}
data augmentation (\eg~cropping, rotating, stretching etc) when keeping a
machine learning model fixed~\citep{
  Salamon2016DeepCN,
  Perez2017TheEO,
  Tsuchiya2019AMO,
  Tian2020WhatMF,
  Laptev2020YouDN}.
Naturally there are also works studying influence of feature selection
methods~\citep{Koak2019InfluenceOS} and projection methods~\citep{Wan2021OnTU}
in addition to those those referenced in the previous paragraphs.

In the realm of classical machine learning, theoretical studies have been
conducted on various models, including decision
trees~\citep{Syrgkanis2020EstimationAI}
and quadratic classifiers~\citep{Latorre2021TheEO}, that explore the estimation
capabilities and the performance within high-dimensional settings.
These models often exhibit a dependence on dimension, particularly in
the context of high-dimensional regimes where their effectiveness may vary.
Similarly, research has been dedicated to understanding the behavior of support
vector machines in spaces with low (box-counting)
dimension~\citep{Hamm2020AdaptiveLR}.

In a different vein, influence function studies track the impact of training
data on learning algorithms, providing insight into how the model's predictions
on test data are influenced by the training data.
This concept has found applications in neural networks, with works shedding
light on the backpropagation process and the attribution of training data
importance in these complex architectures~\citep{
  Koh2017UnderstandingBP,
  Pruthi2020EstimatingTD,
  Akyrek2022TracingKI,
  Hammoudeh2022TrainingDI}.

Otherwise, there seems to be a lack of research on data manipulation methods
that focus on the influence of the concentration of measures phenomenon on model
performance.


\section{An Ontology of Reproducibility in Machine Learning}
\label{sec:ontology}
The term reproducibility is often used ambiguously and vaguely in the field of
machine learning~\citep{
  Peng2011ReproducibleRI,
  Marwick2017AssessingR,
  Plesser2018ReproducibilityVR}.
In our work, we apply the ``classical'' understanding of reproducibility in
science.
That is, whenever a scientific study is replicated the original experimental
results should be archived with a high degree of reliability.
Of course, the concept of ``replication'' implies a series of attributes.
These are essentially related to the fact that it is an independent project,
which could mean a different set of equipment (hardware or software), a
different group of researchers, but could also take into account the time that
has elapsed since the original study.
As such one can see a particular scientific results lying on a spectrum
of reproducibility.
To give a more explicit process of contextualizing reproducibility we introduce
in the following section an extensive \emph{ontology of reproducibility} for the
realm of machine learning.
The necessity for this is based on the fact, that, to the best of our knowledge,
such an ontology does not exist yet.
The components of this ontology are influenced and inspired by the Chapter
\textit{Assessing Reproducibility} of the online version of the book \textit{The
  Practice of Reproducible Research}~\citep{Marwick2017AssessingR}.
Additional ascendancy comes from existing efforts to characterize computational
reproducibility~\citep{
  Gundersen2018OnRA,
  Gundersen2018StateOT,
  Gundersen2022SourcesOI}.
We adapted them with a stronger formalization, giving structure to the
proposed questions and adjusted the ontology to better meet the requirements for
the subsequent reproducibility study.
Other noteworthy influences come from an ontology for semantic terms in machine
learning~\citep{Publio2018MLSchemaET}, a practical taxonomy of machine
learning~\citep{Tatman2018APT} which however has no formalization and very few
specific points to check, and the machine learning reproducibility checklist
~\citep{reproducibility_checklist}.
One commonality between this ontology and the above-mentioned works is the focus
on reproducibility of an individual research project.
We consider only the setting where one computational result is presented as
evidence for one scientific result for simplicity.

\subsection{Overview}
\begin{figure}
  \centering
  \includegraphics[width=0.9\linewidth]{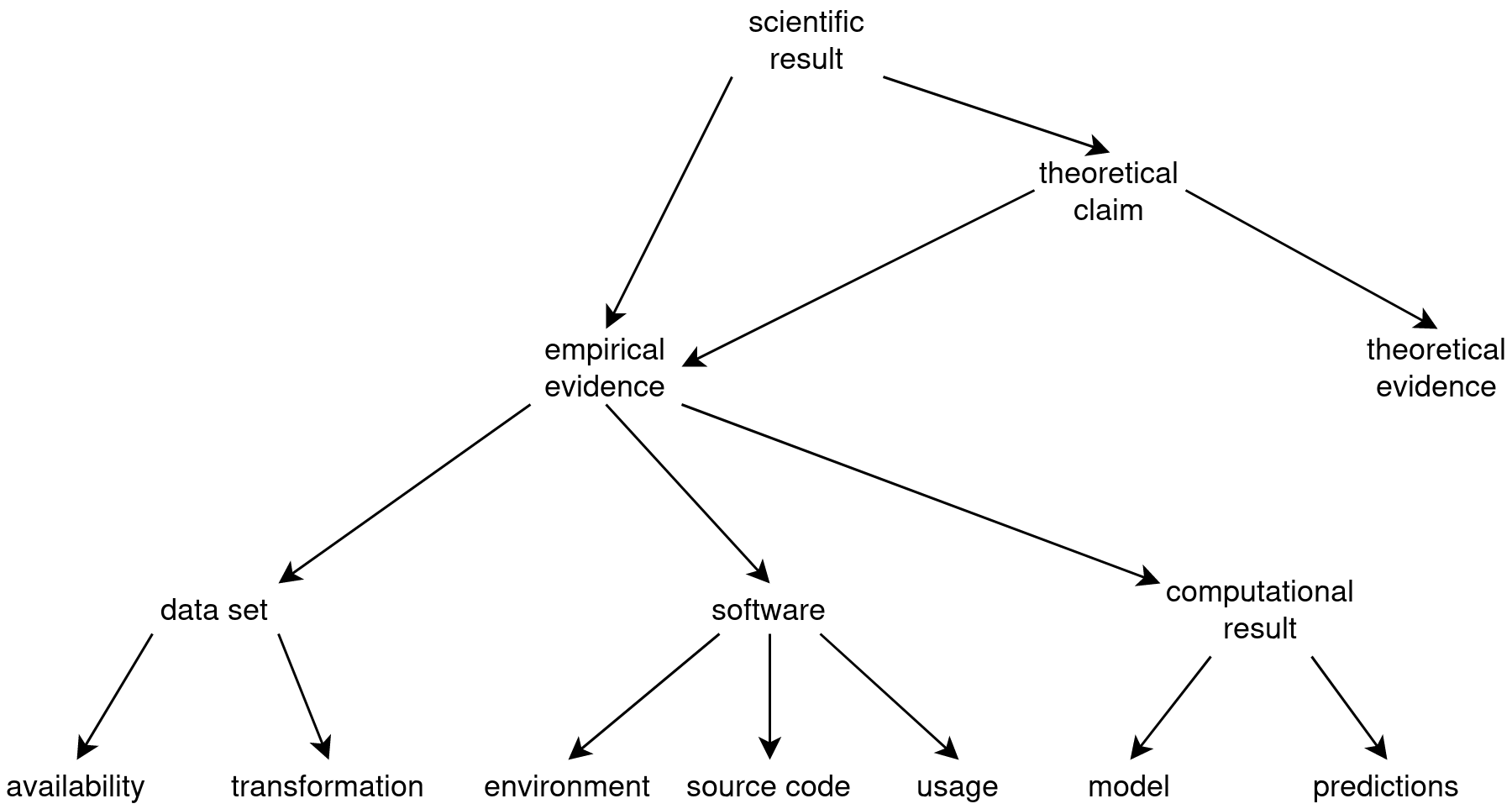}
  \caption{Top levels of the reproducibility ontology.}
  \label{fig:reproducibility_ontology}
\end{figure}

We now present our ontology of reproducibility in machine learning
(\cref{fig:reproducibility_ontology}) which connects possible errors or
difficulties that could arise when trying to reproduce the results of a single
scientific study.
Our proposed ontology is structured as a hierarchy and starts on the top
level with the general notion \emph{scientific result}.
It can be based on \emph{empirical} or \emph{theoretical evidence}.
Because we are (subsequently) interested in research that uses experiments
for producing evidence we do not subdivide the theoretical category.
In contrast we propose a fine-grained structuring for the empirical category.
To reflect the related central aspects of data processing we use the ontological
entities \emph{data set}, \emph{software} and \emph{computational results} as the
main subcategories for the empirical category.
Each of these aggregate again a set of subentities.
For example, the data set category encompasses the \emph{availability} and
\emph{transformation} of data.
Similarily, the software category has as central subcategory \emph{source code}
but also includes \emph{environment} and \emph{usage} as subcategories.
Most importantly, we include the derived \emph{model} and its \emph{predictions}
in the computational result category.
Figure \ref{fig:reproducibility_ontology} shows the main categories and
their subcategories in a overview schema.

We further evaluate every considered research paper within the proposed ontology
based on a set of questions.
In the following we want to describe the formalization of these questions and
their motivations, which are based on potentially occurring errors and their
impact on the scientific reproducibility.
For a detailed list of the ontology we refer to \cref{tab:context:full} in
the \cref{sec:appendix:rep_context}.
As we follow an open world semantic with our ontology the questions are
formulated in such a way that answering them negatively is good for
reproducibility.
This also means that in the subsequent questioning, a missing answer does not
indicate the non-existence of the corresponding property.
We included several questions that remained unanswered for all surveyed
publications (a good thing), but could be helpful in obtaining a finer
distinction of reproducibility if the situation they capture occurs.

\subsection{Data Set}
One central aspect of scientific result in machine learning is the data set to
which previous and proposed methods are applied.
A common approach is to evaluate methods over multiple different but similarly
structured data sets.
Reproducibility is only possible with detailed knowledge about process for
obtaining and preparing used data sets.%
\footnote{But even then there are special cases where methods select data
  samples or generates them (e.g.\ active learning, reinforcement learning), and
  reproducibility is handicapped.}
Those explanation minimize the risk of working with a data that follows a
different data distribution than the one used in the original study.
With this category of the ontology we want to focus on if the publication and
accompanying material include steps (manual or automated) that describe how to
obtain data sets.

\subsubsection{Availability}
The questions from this category aim to reflect nuances in obtaining and
understanding the data sets used by the publication.
\begin{description}
\item[D1\ddash Is the data set format not documented?] Applicable mainly for
  publications that introduce a now or heavily modify an existing data set, it is
  otherwise difficult to adapt methods for reproducibility.
\item[D2\ddash Was the data set version not set explicitly?] Over time data sets
  can undergo changes for individual samples through relabeling or extension,
  which leads to the necessity of keeping track of the used version of the
  data set.
\item[D3\ddash Was the data set not directly accessible?] As most data sets
  currently used in machine learning are distributed over the web, a direct
  download link provides a good start for reproducibility.
\item[D4\ddash Did the access not work at time of study?] Unfortunately the
  provided hyperlinks tend to cease to point to their original resource.
\item[D5\ddash Is the data set privacy restricted?] Although not very frequently
  occurring in broadly used machine learning data sets, license or privacy
  concerns can create restrictions for access.
\item[D6\ddash Does the data set require a restrictive license agreement for
  accessing?]
\item[D7\ddash Is the data set available on request only?] Loosely connected with
  the previous points, if it is necessary to go through a more elaborate process
  for obtaining data set.
\end{description}
\subsubsection{Transformation}
In general a data set needs to be adapted before a method can be applied.
The questions from this category deal with evaluating those pre-processing
steps.
\begin{description}
\item[D8\ddash Are manual steps necessary for pre-processing?] A series of
  manual steps for transforming a data set could easily be a source of mistakes,
  therefor an automatic solution is preferred.
\item[D9\ddash Is there only an incomplete description for pre-processing
  steps?] Furthermore the provided explanations or scripts could not be enough
  to get the data set in the necessary form.
  Especially for lesser known data sets it is helpful if detailed descriptions or
  utility functions are provided that work around intricacies of individual
  samples.
\item[D10\ddash Are the train, validation and test splits unclear?] Usually only
  a part of the data set is used for training, whereas other parts are used for
  validation and testing.
  The process of allocation should be reproducible, be it through provided files
  or deterministic functions.
\item[D11\ddash Is the number of samples not documented?] An easy way of
  checking one attribute of the transformation is counting the obtained samples.
  Additionally it gives a high-level overview over the data efficiency of the
  presented approach.
\end{description}

\subsection{Software}
The implementation and application is a central part of a proposed machine
learning method.
It is one aspect of the research protocol and acts as description of the
executed experiments.
It operates on one or more data sets and produces computational results.
In this category ancillary software and code written by the authors,
as well as reproducibility components connected to hardware, are combined.
This was done to keep the ontology clearly laid out.

\subsubsection{Environment}
Questions from this category deal with general behavior of the target system,
which heavily influences the context of execution of the experiments.

\begin{description}
\item[S1\ddash Is the exact version of dependencies not documented?] Multiple
  dependencies can interact in intricate ways.
  This makes pinning of exact versions necessary for avoiding possible bugs
  connected to incompatible versions as well as prevent time consuming fixing of
  conflicts.
\item[S2\ddash Is the specified version of dependencies not available anymore?]
  Depending on the age of the publication and the type of the used dependency
  the old versions could have disappeared from distribution channels and are not
  hosted by developers or supporters of the project anymore.
\item[S3\ddash Is necessary hardware unavailable?] A lot of special hardware
  requirements can be circumvented by simulations with virtual machine or
  container images, but this leads to time consuming overhead in the
  reproducibility attempt or is not realisable in reasonable time at all.
\item[S4\ddash Are any seeds for random number generators not set?] Multiple
  dependencies each can have different random number generator, where each has
  to be set for getting closer to reproducibility of an experiment.
\item[S5\ddash Are important variables unclear?] Some setting of an experiment
  run (e.g. number of GPUs used) can have significant impact on results or even
  sideeffects onto other settings.
\end{description}

\subsubsection{Usage}
This category of the ontology groups together aspects regarding how the
experiments were started.
For the set of considered machine learning papers it is not necessary to
consider user input beyond starting configuration.

\begin{description}
\item [S6\ddash Is the documentation not up-to-date?] Few publications include a
  dedicated documentation of their provided source code.
  If only a simple \textit{Readme} file is included, it should at least be not
  misleading for the reproducibility attempt with its statements.
\item [S7\ddash Are necessary arguments not clear?] Depending on implementation
  some arguments might me necessary to run an experiment but neither defaults or
  used values are explained or provided.
\item [S8\ddash Are there missing hyperparameters?] Similar to previous question
  the values for hyperparameters for the experiments are generally of importance
  when reproducing it as they influence outcome significantly.
\item [S9\ddash Are train/test scripts incomplete?] Including the individual
  commands of an experiment in a single file usually facilitates the
  reproducibility attempt.
  How to start these scripts can be a source of uncertainty if certain flags or
  variable values used in provided script are missing, wrong, only corrected
  later and/or not explained at all.
  Additionally if pre-processing steps are not included or other steps in the
  computational pipeline are missing, reproducibility is affected.
  Furthermore it could be the case that not all experiments presented in the
  paper have scripts.
\item [S10\ddash Is it unclear which version of scripts was used?] It is only
  natural for the main source code of the publication to undergo changes before
  (and after) publication to accommodate for bugs and reviews.
  Problems arise when those changes are not reflected in accompanying scripts or
  instructions or when multiple scripts for same experiment exist
  simultaneously.
\end{description}

\subsubsection{Source Code}
The questions in this category are concerned with the source code files that
implement the ideas of the scientific experiment.
We do not have separate questions regarding the availability of source code
itself similar to the data set category because we only focus on papers that
provide source code with a non-restrictive license.
\begin{description}
\item [S11\ddash Is there a bug that was never fixed?]
  Over time authors and external contributors could find differences or errors
  between original publication, its revisions and the implementation.
\item [S12\ddash Are there issue solutions that were not applied?]
  Usually the detection of problems of the implementation is accompanied by
  public discussions on the website that hosts the implementation.
  But it can happen, that the discussed solution was neither implemented or
  merged from external source code fork.
\item [S13\ddash Was a bug fix distributed through other channels?]
  On the other hand one can get a hint in public discussions that the fix was
  distributed (manually) through some other channel like emails or direct
  messages to selected/active group of participants.
  In those cases it is not clear what the detailed changes where and how they
  affect reproducibility.
\item [S14\ddash Did the API change?]
  This question is connected to an above usage question regarding different
  versions of scripts.
  Now we consider the other side, e.g.\ the entry points to core parts of the
  implementations changed but this is not reflected elsewhere be it
  documentation or other supporting material.
  The reproducibility effort is furthermore increased if traceability of
  versions is limited by convoluted history in the version control system.
\item [S15\ddash Did an out of memory error occur?]
  Considered as a special type of error it is only recorded if there are no
  specification from requirements or those are not correct.
  As our goal is to evaluate reproducibility generally we do not determine the
  specifics that caused this error.
\item [S16\ddash Are steps for one experiment missing?]
  If necessary source code for one experiment is not included, reproducibility
  for this experiment can only obtained with more difficulties.
\item [S17\ddash Are steps for all experiments missing?]
  Additionally to the previous point we want to evaluate the possible situation
  that the publication uses libraries or code that is not included in the
  provided source code.
  This makes reproducibility nearly impossible.
\item [S18\ddash Is the hyperparameter search not included?]
  As hyperparameter search is integral part of experiments it is important for
  reproducibility to have an explanation or process in the implementation on how
  the search was carried out.
\item [S19\ddash Is only the general idea (and no experiments) implemented?]
  Another reason for the absence of experiments could be that the publication
  only proposes a new machine learning algorithm or a building block for an
  existing one.
\end{description}

\subsection{Computational Result}
The result obtained through a computational experiment represent the evidence
of a scientific claim.
A full reproduction of a scientific result, and the corresponding evaluation,
depends on the successful completion of the reproducibility steps for data set and
software.
Problems arise when this is not the case.
This implies that we can not obtain a model or predictions for the further
evaluation steps.
If the supplementary material of the publication in question does
include the learned model it is still possible to perform the subsequent
reproducibility steps.
However, this is a rare case.
Ideally contemporary scientific results in machine learning should be
reproducible in terms of data set and software as well as provide the learned
model.
This case even allows for a more in-depth comparison and analysis between the
reproduced and the provided models.

\subsubsection{Model}
For now there is only one questions in this category.
As outline above the simple access to model weights is necessary for full
reproducibility, especially if other factors increase difficulty of obtaining
a optimized model independently.
As such it is often overlooked but even when considered, the practical problem
of making the model available has still no ready-made solution.
There are a few existing platforms that allow for combined hosting of source
code, data sets and models.
Limitations can be encountered quickly if data sets and models are large or
several of them are used or provided.

\begin{description}
\item [R1\ddash Are there no parameters (weights) of the obtained model
  provided?] As large parts of contemporary machine learning approaches use
  larger data set and models, it takes more and more time and computing
  resources to run the proposed method.
  Providing model weights can therefor act as a form of shortcut if comparison
  with other approaches is in focus.
  Depending on programming language and format it provides checks on
  specifications of models.
  Additionally when reproducibility fails it gives possibility to find cause
  and more importantly at least check author claims that way.
\end{description}

\subsubsection{Predictions}
With this category we want to capture aspects of the output of the
train/test data set of the model and how these affect reproducibility.
Depending on presented results a comparison of a variety of evaluation metrics
can be helpful, especially when inference takes longer time or larger resources
requirements than available for the specific reproducibility attempt.
\begin{description}
\item [R2\ddash Are there small deviation to obtained model?]
  We answer this questions positively when comparing the central evaluation
  metrics reported by original authors with evaluated metrics on the reproduced
  model a difference in the range of more than $\pm 1-2\%$ is observed.
\item [R3\ddash Are strong differences in few experiments observed?]
  Similarly to above we assign this attribute when the difference of evaluation
  metrics are in a range of more than $\pm 10-20\%$.
\item [R4\ddash Are strong differences in almost all experiments observed?]
  As an extension to the previous question we assign this attribute when almost
  no reasonable reproducibility of outcomes can be obtained.
\item [R5\ddash Are the claimed results only supported by small sample size?]
  Individual runs of a machine learning algorithm are rarely exact reproducible,
  even with the best efforts for obtaining reproducibility, by both original
  authors and those who reproduce the work.
  By averaging over a few executions the expressiveness of the results can be
  strengthened.
  We set the threshold for this at less than 5 samples.
\item [R6\ddash Are there no predictions (outputs of classes or decisions) on
  the data sets?]
  If the original publication provides the class predictions or decisions made
  otherwise by the model it gives the reproducibility attempt the possibility to
  investigate more comprehensive metrics for differences in model behavior.
  Although making the complete set of predictions available is not always
  feasible (e.g.\ for large data sets or methods from other fields such as
  reinforcement learning), it could be for non-trivial parts of data set.
\end{description}

\subsection{Limitations and Extensions}
It is apparent that this ontology is designed using a basic formalization language.
All connections between the entities can be read as \emph{part of}.
The authors are well-aware of the \textit{ML-Schema}~\citep{Publio2018MLSchemaET}.
However, we decided to not include or build upon it, since:
i) several aspects of reproducibility could not be expressed using ML-Schema;
ii) the focus of the ML-Schema is on interchanging information on machine learning
algorithms and not on the reproducibility of an scientific result.
An example for this is the lack of consideration of the influence of the seed
for the random number generator.

There is a series of possible changes or extensions that we did
not include in the presented version of the ontology.
For example, our ontology does not consider detailed information about
theoretical evidence.
This is mainly motivated by the survey in \cref{sec:survey} that focuses on
empirical results.
In the empirical evidence category the designation \emph{software} is
slightly misleading because it additionally contains aspects related to
hardware.
One may rename this category or add \emph{hardware} as its own subcategory of
empirical evidence.
Analogously the entity \emph{source code} might be extended with
details about different modes of availablity and documentation.
A special attribute might be the application of version control software.

Certain aspects of reproducibility are not considered yet, e.g.\ that
plots, figures and tables can be automatically generated.
Often authors omit to provide the corresponding source code for visualization.
Furthermore, our ontology does not capture data provenance aspects.

Additionally it seems the longer it has been since publication, the lower the
achievable degree of reproducibility.
This can be exemplified by aging hard- and software that was used for the
experiment and might not be available anymore.

Finally in our ontology we treat the presence or absence of an attribute
categorically.
Hence in certain cases evaluating a research work our ontology is a difficult
task.
Conversely any subsequent ontological operation and explanation is independent
of any interpretation of numerical values.


\section{Reproducibility of Major Graph Neural Network Research Results}
\label{sec:survey}
The first main goal of the present work is to achieve a scientific overview
over the state of reproducibility in the research field of graph neural networks.
For this we first depict our method of candidate selection in
\cref{sec:candidate_selection} and thereafter discuss all our findings with
respect to our reproducibility ontology.

\subsection{Candidate Selection}
\label{sec:candidate_selection}
The main criterion for selecting a paper was its impact on the research field of
graph neural networks.
In the following we describe in detail our procedural steps.
As a lower bound for the publication year we selected 2016, the year of the
publication of the seminal \textit{GCN} paper~\citep{Kipf2016SemiSupervisedCW}.
On the other hand we considered works that were published before 2023.
Most importantly we required that the paper in question has an experimental
evaluation, due to the overall objective of the study to investigate the
influence of intrinsic dimension.
We also included research works that were only in the preprint stage.

As the citation count is an often used proxy for measuring scientific
impact, and at the same time readily available, we employ it in our selection
process.
In detail, we use the \textit{Semantic Scholar}~\citep{semantic_scholar} search
engine for selecting papers based on their average citation count since their
publication.%
\footnote{In other words, we rank search results for the keyword \textit{graph
    neural network} based on the score: $\frac{\text{number of
      citations}}{2023-\text{year of publication}}$.}
In our selection process we discarded all papers without publicly available
source code for their experiments.
We further discarded papers that either covered implementation details of
software libraries or focused on applications of existing methods.
Finally we refined our selection with the help of several domain experts that
pointed us to important research works from the domain of graph neural networks.

Our selection process started out with \input{results/unique.dat}\unskip
unique candidates.
We then calculated each paper's score and selected the
\input{results/relevant.dat}\unskip papers with the highest scores.
We manually ignored works that did not propose a new method in the field of
machine learning.
This included surveys, coding frameworks, and works that only applied Graph
Convolutional Network (GCN) methods to other field of science.
Additionally, publications applying methods to very specific data sets and those
with time-dependent or spatial data were not included.

\begin{figure}[h]
  \begin{subfigure}{.5\textwidth}
    \centering
    \includegraphics[width=\linewidth]{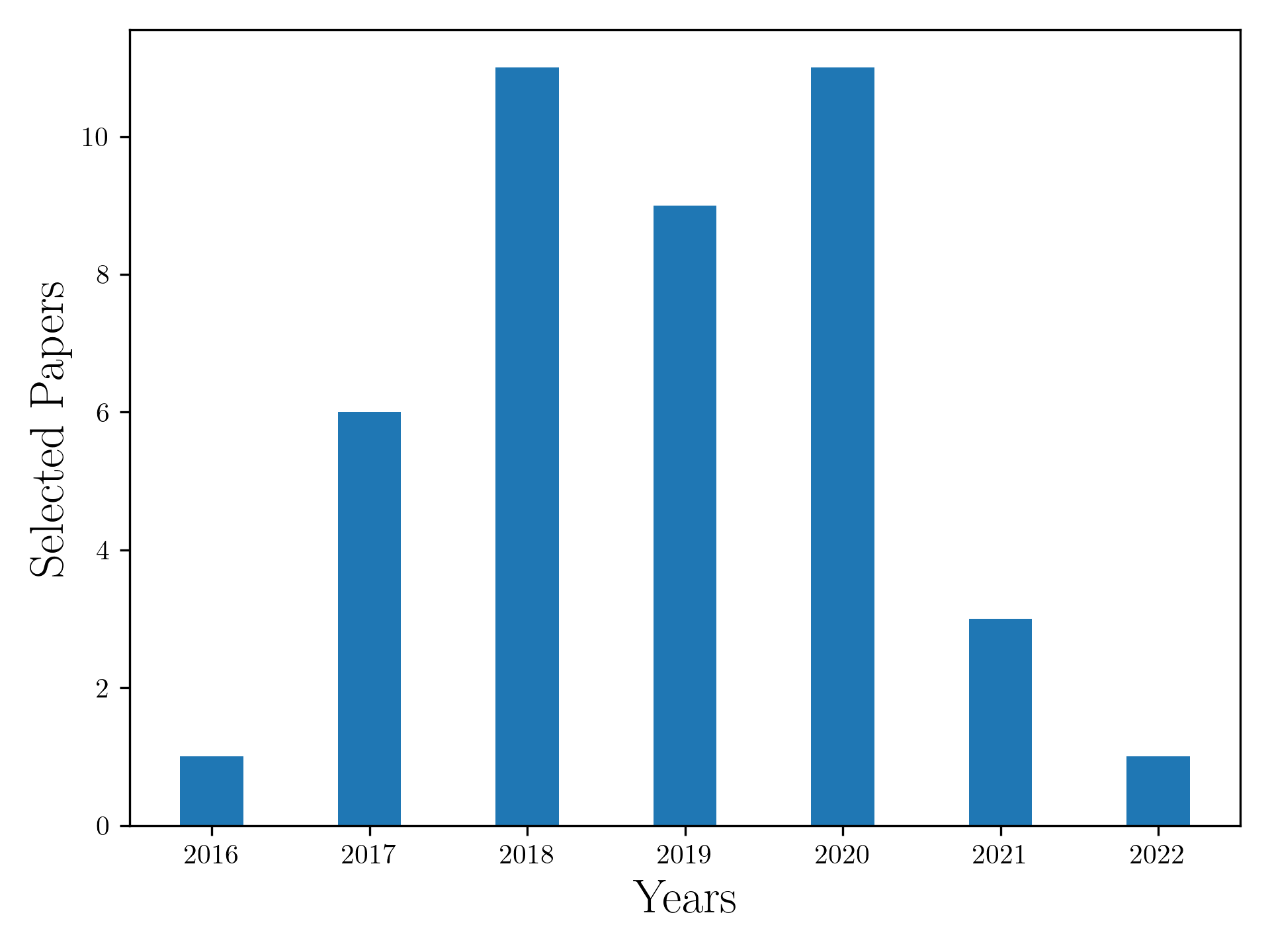}
    \caption{Number of selected papers per year.}
    \label{fig:papers_history}
  \end{subfigure}
  \begin{subfigure}{.5\textwidth}
    \centering
    \includegraphics[width=\linewidth]{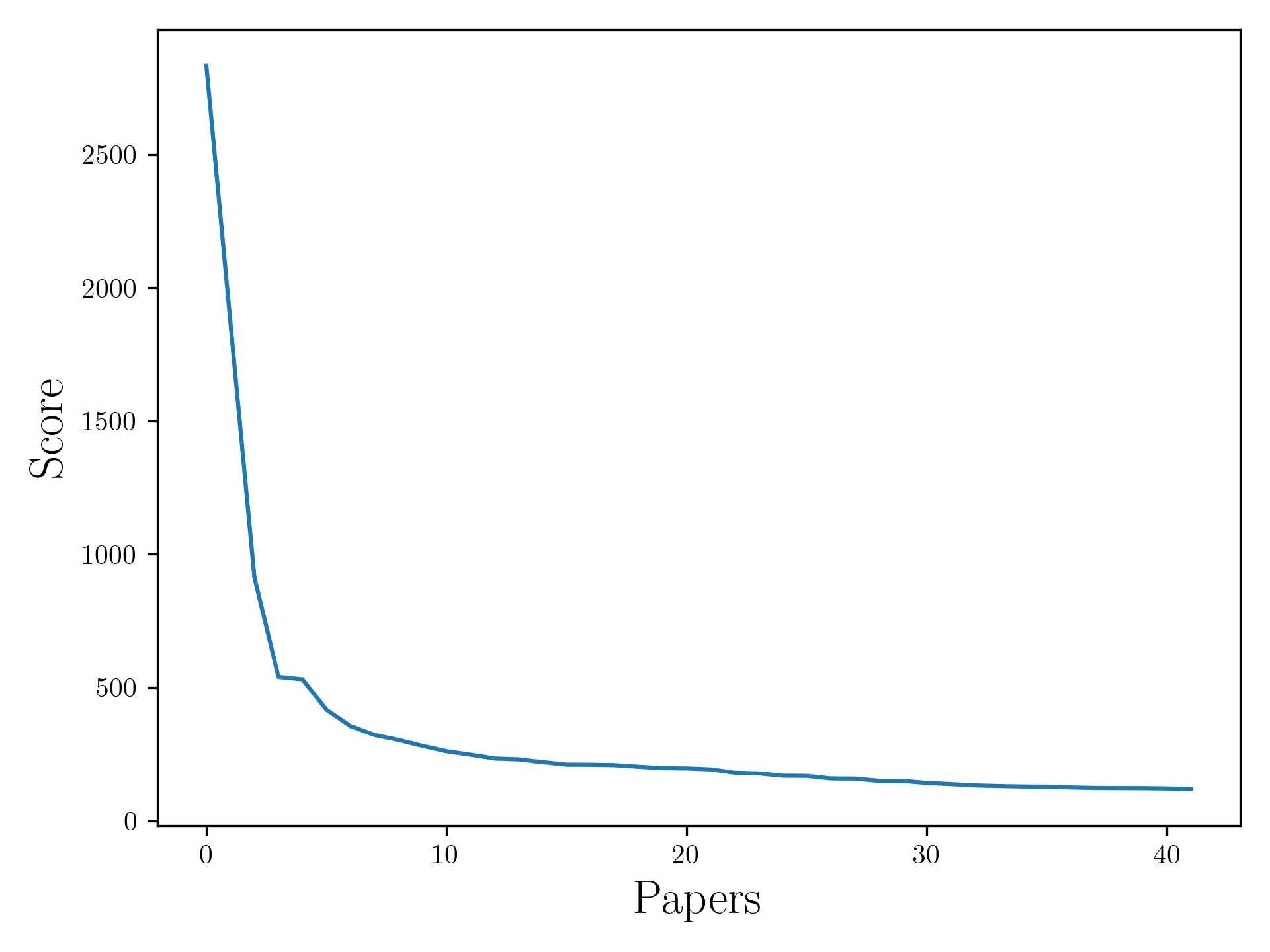}
    \caption{Selected papers sorted by score.}
    \label{fig:papers_scores}
  \end{subfigure}
  \caption{Visualization to get an overview of the distribution of the
    selected papers.}
  \label{fig:papers_distribution}
\end{figure}

Out of those remaining \input{results/investigated.dat}\unskip results we
applied the source code criterion and arrived at
\input{results/with\_code.dat}\unskip papers.
In \cref{fig:papers_distribution} we depict the yearly distribution of the
number of papers (a) and their score distribution (b).

Now the following limitation of the selection method becomes more apparent.
As citations are distributed over publications and there is an increasing number
of papers published each year, older papers have advantage over newer ones.
Conversely, the evaluation function dampens the influence of older publications
to a much lesser extent.

On the other hand we wanted our selection method to reflect a ``normal'' search
behavior of a researcher.
The power-law distribution of citations is a well known property of citation
networks~\citep{Price1965NetworksOS} and also somewhat expected because they are
social networks which accompany the scientific process.
More specifically methods of more frequently cited papers are chosen more often
than those with less citations~\citep{Hazoglu2017CitationHO}.
The power-law distribution of the citations (and subsequently the score)
motivated selecting only a few papers as those publications
had the majority of the impact on the field measured by the above method.

We provide a list of considered papers in \cref{tab:survey:considered} in the
appendix.
We starting with selecting candidates for the reproducibility survey before
deciding on the automated process presented above.

Incompatibility in hardware requirements and bugs within the provided source
code were the primary reasons for determining at this stage if a paper was not
reproducible.
If these issues could not be resolved with reasonable effort, even with
experience using the libraries, the paper was deemed irreproducible.
Regrettably we were not able to fully map the preceding manual pre-selection to
an automated process.
The difficulty in reproducing the results was due to the uncontrollable
non-determinism introduced by the usage of the \emph{Semantic Scholar API} over
multiple runs and new citation data resulting in changes in the rankings over
several months.

Furthermore, the collections used did not always include papers for which the
reproducibility attempt failed.
However, once we obtained a set of considered papers, we refrained from further
optimizing the selection process to include all reproduced or not reproduced
papers.
We acknowledge the possibility of producing a positivity bias by excluding
publications where reproducibility failed completely.
However, we believe this is preferable to conveying a skewed perspective of the
reproducibility of the survey itself and the field of graph neural networks in
general.

Therefore, the collection contains \input{results/reproduced.dat}\unskip
publications, for which we completed the complete assignment of the described
reproducibility attributes.
\emph{SGC} and \emph{GraphSAGE} were successfully reproduced candidates
determined from prior iterations of the selection process.
Since we already had experience with the publication for \emph{SAGN+SLE}, we
included it as well.
The final selection of reproduced papers can be seen in
\cref{tab:survey:papers} together with their abbreviations used in the
following.

\newcommand{\GCN}{GCN}
\newcommand{\RGCN}{R-GCN}
\newcommand{\GraphSage}{GraphSAGE}
\newcommand{\DiffPool}{DiffPool}
\newcommand{\SGC}{SGC}
\newcommand{\SAGNSLE}{SAGN+SLE}
\begin{table}[h]
  \begin{center}
    \caption{Reproduced Papers and the abbreviations used.}
    \begin{tabular}{|c|p{9cm}|c|}
      \hline
      Abbreviation & Title & Reference Key \\
      \hline
      \GCN & \citetitle{Kipf2016SemiSupervisedCW} & \cite{Kipf2016SemiSupervisedCW} \\
      \hline
      \RGCN & \citetitle{Schlichtkrull2017ModelingRD} & \cite{Schlichtkrull2017ModelingRD} \\
      \hline
      \GraphSage & \citetitle{Hamilton2017InductiveRL} & \cite{Hamilton2017InductiveRL} \\
      \hline
      \DiffPool & \citetitle{Ying2018HierarchicalGR} & \cite{Ying2018HierarchicalGR} \\
      \hline
      \SGC & \citetitle{sgc_proceedings} & \cite{sgc_proceedings} \\
      \hline
      \SAGNSLE & \citetitle{sagn_sle_arxiv} & \cite{sagn_sle_arxiv} \\
      \hline
    \end{tabular}
  \end{center}
  \label{tab:survey:papers}
\end{table}

\subsection{General Observations with Respect to our Ontology}
Reproducing experimental results from the selected scientific research was
challenging due to various factors.
It is usually the case that papers taken alone do not provide enough
information to replicate the experiments independently.
Correspondingly we have always started the reproducibility attempt with the
associated source code.
One major issue is that those repositories often lack crucial dependency
information (S1), making it difficult to even run the entry point scripts without
errors.
To overcome these challenges, it was necessary to search through accompanying
discussions and seek clarification on exact parameters and commands that may be
missing or not functioning correctly.
Additionally it is often the case that the commands provided in the simple
documentations to run experiments rarely work as expected (S6, S7).
Furthermore, the availability of different data sets adds complexity to the
reproducibility process, especially considering that many publications were
released before coordinated efforts to unify the data set landscape, for example
the \textit{Open Graph Benchmark}~\citep{Hu2020OpenGB}.
However, data sets are generally accessible, although it is rarely the case that
the preprocessing steps are explained (D9).
Another common point is the aspect of hyperparameter search, which is not
included in most provided software, even when mentioned in the publication
(S18).
Lastly, it is common for papers to lack the provision of the model (R1) or
predictions of the model on specific data (R6).
We will include in the following a more detailed description of specific problems
grouped by the main categories \emph{Data Set, Software} and \emph{Computational
  Result} that were encountered during the reproducibility attempts.
For a better overview we will focus on selected points that stand out.

We want to emphasize that in our ontology it is \textit{not} beneficial to have
an attribute as it is evidence that the reproducibility is more difficult.
In cases where it was not clear whether an attribute was present, we chose not
to disclose it.

\renewcommand{\longtextorientation}{0}
\explanationfalse

\newcommand{\pfill}{:\quad}

\subsubsection{Category: Data Set}
\label{sec:candidate_selection:obs:data}
\subsubsubsection{\GCN}\pfill
We observe that the data set used in the research is conveniently available as it
is included in the repository.
However, there is a lack of explanation regarding the preprocessing steps (D9).
Despite this, a working function is provided, which can be used for the
transformation process.

\subsubsubsection{\RGCN}\pfill
A similar point as in \GCN~regarding (D9) applies.

\subsubsubsection{\GraphSage}\pfill
The availability of the \textit{web of science} data set is limited to those
with the corresponding license (D6) and upon request (D7).
Additionally, manual preparation (download) of the data sets is required (D8).

\subsubsubsection{\DiffPool}\pfill
The implementation does not use proper train/test splits because the approach is
only evaluated with k-fold validation (D10).

\subsubsubsection{\SGC}\pfill
A similar point as in \GCN~regarding (D9) applies.

\subsubsubsection{\SAGNSLE}\pfill
Except of the general observations of missing explanation of the preprocessing
steps (D9) the aspects of the data set category are sufficiently reproducible.

\begin{table}[h]\label{tab:survey:data set}
  \caption{Observations for \emph{data set} category with respect to
    \ref{sec:candidate_selection:obs:data}.}
  \datasettable{\tiny}
\end{table}

\subsubsection{Category: Software}
\label{sec:candidate_selection:obs:software}
\subsubsubsection{\GCN}\pfill
The dependencies are not properly specified (S1), which made it challenging
to set up and run the experiments.
It is worth noting that the documentation is not up to date (S6) and contains
misleading information.

\subsubsubsection{\RGCN}\pfill
Firstly, there is no requirements file provided (S1), making it challenging to
recreate the necessary environment.
Additionally, information about the specific Python interpreter version used is
hidden.
Furthermore, the seeds for randomization are not set (S4).

\subsubsubsection{\GraphSage}\pfill
The necessary arguments for the evaluation scripts are not stated (S7), leaving
researchers unsure of the required inputs.
Furthermore, there are discussions about possible values, adding ambiguity to
the code (S8).
Additionally, the evaluation scripts themselves are incomplete or misleading,
further hindering reproducibility (S9).
The software used in the study has some bugs that affect reproducibility (S11).
For example, the evaluation script for the \textit{ppi} data set is incomplete,
but a fix is available in pull requests (S12).

\subsubsubsection{\DiffPool}\pfill
Unfortunately there is no explicit list of necessary requirements (S1) and only
a minimal README file (S6).
Additionally it seems that the provided commands do not work (S7, S8 and S9) and
that the seeds for randomization are not set before the experiments (S4).
The implementation also did not include steps to reproduce two experiments with
the \textit{reddit-12k} or \textit{collab} data sets (S16).

\subsubsubsection{\SGC}\pfill
No features of that category that hindered the reproducibility were observed.

\subsubsubsection{\SAGNSLE}\pfill
Again, there is no requirements file provided (S1).
Unfortunately we encountered out-of-memory errors (S15) when trying to reproduce
experiments using data sets \textit{ogb-papers} and \textit{ogb-mag}.
It could be argued that this would mean that the necessary hardware is
unavailable (S3) but maybe it could be fixed by changing hyperparameters like
batch size.

\begin{table}[h]\label{tab:survey:software}
  \caption{Observations for \emph{software} category with respect to
    \ref{sec:candidate_selection:obs:software}.}
  \softwaretable{\tiny}
\end{table}

\subsubsection{Category: Computational Result}
\label{sec:candidate_selection:obs:results}
\subsubsubsection{\GCN}\pfill
When examining the results, it is observed that there are small deviations in
the test set accuracy, typically within a range of $\pm 1\%$ (R2).
However, the statistical analysis in the paper is weak as it does not provide
information on the standard deviation (R5).
Although the authors claim to have run the experiments with multiple seeds,
there is no evidence of this in the code, which raises concerns about the
robustness of the reported results.
We were not able to reproduce the experiments with the neil data set because of
the difficulties to prepare and use the data set.

\subsubsubsection{\RGCN}\pfill
A similar observation regarding multiple runs was made for this publication as
well.
Additionally, the results of the study exhibit both small deviations and strong
differences in different parts.
For the AIFB, MUTAG, and AM data sets, small deviations of approximately $\pm
2\%$ in accuracy are observed (R2).
However, for the BGS data set, a significant difference of $15\%$ is observed,
indicating a substantial variation in the results (R4).

\subsubsubsection{\GraphSage}\pfill
The results of the study exhibit small deviations, typically within a range of
$\pm 2\%$, for the available data sets (R2).
However, the statistical analysis is weak, suggesting that the experiments were
only run once (R5).

\subsubsubsection{\DiffPool}\pfill
Due to time constraints we were only able to reproduce the experiments for the
DD and Enzymes data set and observed small deviations $\sim 2\%$ (R2).
Even though k-fold validation was used the experiment was only run once (R5).

\subsubsubsection{\SGC}\pfill
No feature other than the usual were observed.

\subsubsubsection{\SAGNSLE}\pfill
Similar to other survey candidates we obtain results that exhibit small
deviations, typically within a range of $\pm 2\%$, for the available data sets
(R2).

\begin{table}[h]\label{tab:survey:results}
  \caption{Observations for \emph{computational result} category with respect to
    \ref{sec:candidate_selection:obs:results}.}
  \resultstable{\tiny}
\end{table}

\subsection{Discussion}
\label{sec:survey:discussion}
Regarding the reproducibility ontology we observed that most paper look almost
the same in \emph{data set} category but needed quite different effort during
reproducibility attempt.
This is particularly visible in the fact that attributes D1 to D5 are not
fulfilled in any of the attempts.
The ease of reproducibility was decided mainly through information
provided in the README document of the source code.
This means that the ontology does not capture this aspect that well for this
category even if there is a corresponding attribute in the source code category.

The \emph{software} category on the other hand allowed for a very good
differentiation of the different papers with regard to their reproducibility.
Here too, some attributes do not seem to be contribute to deciding on the degree
of reproducibility.
However, further reproducibility attempts may find that these currently unused
attributes become helpful for this goal.

The \emph{computational result} category has some attributes that are common to
all reproduced papers.
This suggests that the corresponding properties (providing model weights and
predictions) are the most difficult to obtain.
We observed multiple papers that included only a low number of repetition for
the experiment in the original paper.
This could be due to higher computing requirements.
We also found that we had included too few cases of possible evaluation
scenarios, and that those that were included were too broadly defined.
Furthermore, there are no attributes to assess the degree of reproducibility of
follow-up or downstream tasks also addressed in the original work.


\section{Influence of Intrinsic Dimensionality on Model Performance}
\label{sec:dimensionality}
The second main goal of the present work is to investigate the influence of
intrinsic dimensionality on model behavior.
We begin by stating the mathematical groundwork of the concept of geometric
\ac{ID} in~\Cref{sec:dimensionality:foundations} and afterwards
present our experiments and results.

As already mentioned the geometric intrinsic
dimension~\citep{Hanika2022IntrinsicDO} is a computational accessible approach
for measuring how a given data set is affected by the \textit{phenomenon of
  concentration of measure}~\citep{Gromov1983ATA, Milman1988TheHO,
  Milman2000TopicsIA}, which itself is deeply connected to the
\cod~\citep{Pestov1999OnTG, Pestov2007IntrinsicDO, Pestov2007AnAA,
  Pestov2010IntrinsicD, Pestov2010IndexabilityCA}.
Of central importance are feature functions that \textit{concentrate}.
This means that they map most of the values of their domain near the mean or
median of their image set.
Pestov has surmised that features of this type contribute the most to the \cod.
In his approach all 1-Lipschitz function are considered as potential feature
functions.
In the revised axiomatic system introduced by~\cite{Hanika2022IntrinsicDO} the
notion of a \textit{dimension function} emerged.
Such a function allows for estimating the extent to which the provided function
concentrate on the data set without having to evaluate all possible feature
functions.
This is motivated by the fact that machine learning algorithms usually
only have access to a limited selection of feature functions of this type.
The computations or approximations of the dimension function of a data set were
improved in recent works~\citep{Stubbemann2022IntrinsicDF}.
In this work we want to build upon the results obtained by using the geometric
intrinsic dimension for feature selection~\citep{Stubbemann2023SelectingFB}.

\subsection{Foundations of the Concentration-based Intrinsic Dimension}
\label{sec:dimensionality:foundations}
We start by briefly recapitulating the mathematical definitions that the
\acl{ID} builds upon.
The interested reader is referred to the cited works for more in depth
explanations.

\begin{definition}
  \label{def:geo-data-set}
  \normalfont\ Let $\mathcal{D}=(X, F, \mu)$ be a triple consisting of a set $X$
  of \textit{data points} and a set $F\subseteq\mathbb{R}^X$ of \textit{feature
    functions} from $X$ to $\mathbb{R}$.
  Consider the function $d_F(x,y)\coloneqq\sup_{f\in F}|f(x)-f(y)|$.
  We require that $X$ fulfills $\sup_{x,y\in X} d_F(x,y)<\infty$ and
  $(X,d_F)$ is a complete and separable metric space with $\mu$ being a Borel
  probability measure on $(X,d_F)$.
  We call $\mathcal{D}$ a \textit{geometric data set}.
\end{definition}

In the following we will limit our considerations to the special case of
\textit{finite geometric data sets}, hence those with $0<|X|,|F|<\infty$ and
$\mu$ being the normalized counting measure as a further restriction.

We will now introduce the building blocks that give rise to a dimension
function that fulfills the aforementioned axioms postulated in
\cite{Hanika2022IntrinsicDO}.
Such a function will indicate a geometric data set with data points that can be
better discriminated by the corresponding set of feature functions by a low
value.

\newcommand{\PD}{\operatorname{PartialDiameter}}
\newcommand{\OD}{\operatorname{ObservableDiameter}}
\newcommand{\del}{\, \mathrm{d}}

Given a feature $f\in F$ we want to evaluate how it can discriminate sets of a
specific measure (e.g. size $c_\alpha\coloneqq\lceil |X|(1-\alpha)\rceil$) for a
fraction $\alpha\in (0,1)$ of the whole $X$.
For this we use can use the following function:
\begin{equation}
  \PD{(f, 1-\alpha)}_{\mathcal{D}} =
  \mathop{\min_{M\subseteq X}}_{|M|=c_\alpha} \max_{x,y\in M} |f(x)-f(y)|.
\end{equation}
By considering all feature from the feature set $F$ we arrive at the
\begin{equation}
  \OD(\mathcal{D}, -\alpha)\coloneqq\sup_{f\in F}\PD(f, 1-\alpha)_\mathcal{D}.
\end{equation}
When considering all possible values for $\alpha$ we obtain a way to describe
the ability of a feature set $F$ of a geometric data set to discriminate data
points in $X$:
\begin{equation}
  \Delta(\mathcal{D})\coloneqq\int_0^1 \OD(\mathcal{D}, -\alpha) \del\alpha
\end{equation}
It turns out that we need one more step to get the \textit{dimension function}
we are looking for:
\begin{equation}
  \partial (\mathcal{D})\coloneqq \frac{1}{\Delta(\mathcal{D})^2}
\end{equation}

For the case of finite geometric data sets it follows that the \ac{ID} can be
explicitly calculated with the help of the following expression
\begin{equation}
  \Delta(\mathcal{D}) = \frac{1}{|X|}\sum_{k=2}^{|X|}\max_{f \in F}
  \mathop{\min_{M\subseteq X}}_{|M|=k}\max_{x,y \in M}|f(x) - f(y)|.
\end{equation}
Using the notation  $\phi_{k,f}(\mathcal{D})\coloneqq \min_{M \subseteq
  X,|M|=k}\max_{x,y \in M}|f(x) - f(y)|$, and $\phi_k(\mathcal{D}) \coloneqq \max_{f \in F}
\phi_{k,f}$ this can be
rewritten as
\begin{equation}
  \Delta(\mathcal{D})=\frac{1}{|X|}\sum_{k=2}^{|X|} \max_{f\in F}
  \phi_{k,f}(\mathcal{D})= \frac{1}{|X|}\sum_{k=2}^{|X|} \phi_k(\mathcal{D}).
\end{equation}

\subsubsection{Approximation of Intrinsic Dimension}
\label{sec:dimensionality:approximation-intrinsic-dimension}
The straightforward computation of the equations in the previous section is
hindered by the task to iterate through all subsets $M \subseteq X$ of size $k$.
This yields an exponential complexity with respect to $|X|$ for computing
$\Delta(\mathcal{D})$.
As suggested by~\cite{Hanika2022IntrinsicDO} and later proven
by~\cite{Stubbemann2022IntrinsicDF}, we can instead use algorithms with a
quadratic runtime complexity in $|X|$ to compute the \ac{ID}.
Furthermore for settings where a quadratic runtime is still not sufficient,
the authors propose the following concept.

Let $s=(2=s_1, \dots,s_{l-1},s_l=|X|)$ be a strictly increasing and finite
sequence of natural numbers.
We call $s$ a \emph{support sequence} of $\mathcal{D}$.
We additionally define
\begin{align}
  \begin{split}
    \Delta(\mathcal{D})_{s,-}&\coloneqq \frac{1}{|X|}
                               \left(\sum_{i=1}^l\phi_{s_i}(\mathcal{D}) +
                               \sum_{i=1}^{l-1}\sum_{s_i<j<s_{i+1}}\phi_{s_i}(\mathcal{D})\right),\\
    \Delta(\mathcal{D})_{s,+}&\coloneqq \frac{1}{|X|}
                               \left( \sum_{i=1}^l\phi_{s_i}(\mathcal{D}) +
                               \sum_{i=1}^{l-1}\sum_{s_i<j<s_{i+1}}\phi_{s_{i+1}}(\mathcal{D})\right)
  \end{split}
\end{align}
and call accordingly $\partial(\mathcal{D})_{s,-}\coloneqq
\frac{1}{\Delta{(\mathcal{D})_{s,+}}^2}$ the \emph{lower intrinsic
  dimension} of $\mathcal{D}$ and $\partial(\mathcal{D})_{s,+}\coloneqq
\frac{1}{\Delta{(\mathcal{D})_{s,-}}^2}$ the \emph{upper intrinsic
  dimension} of $D$.

This results in giving us lower and upper bounds for $\Delta(\mathcal{D})$ and
thus for the \ac{ID}.
By using upper and lower bounds, we can obtain the following approximation of
the \ac{ID}:
\begin{equation}
  \partial(\mathcal{D})\simeq \partial(\mathcal{D})_{s}\coloneqq
  \frac{\partial(\mathcal{D})_{s,+} + \partial(\mathcal{D})_{s,-}}{2}.
\end{equation}

~\cite{Stubbemann2022IntrinsicDF} provides an algorithm for calculating this
approximation.

\subsection{Dimension based Feature Selection}
The intrinsic dimension of a data set refers to a measure of concentration that
capture the underlying structure or information of the data.
It is challenging to quantify the impact of intrinsic dimensionality on a
particular machine learning method, which motivates the need to investigate its
effect.
On way to achieve that is by discarding the features that have the most
significant influence on the dimensionality of the data set.
By removing these features, we can observe whether there is a change in the
performance of the trained model or not.
This approach allows us to examine the relationship between intrinsic
dimensionality and model performance.
Feature selection can be seen as a means to an end in this research.
It serves as a tool to identify and eliminate the features that contribute the
most to the dimensionality of the data set.
To calculate the influence of dimensionality and perform feature selection, we
rely on methods demonstrated in~\cite{Stubbemann2023SelectingFB} which we
will briefly include in the following.

The \emph{discriminability of $\mathcal{D}$ with respect to feature $f
  \in F$} is defined as
\begin{equation}
  {\Delta{(\mathcal{D})}_f^{*}} \coloneqq
  \frac{1}{\lvert X \rvert}\sum_{k=2}^{\lvert X \rvert} \phi_{k,f}(\mathcal{D}).
\end{equation}

Note, that one data point with an outstanding value $f(x)$ can have a strong
influence on ${\Delta{(\mathcal{D})}_f^{*}}$ via drastically increasing
$\phi_{\lvert X \rvert,f}(\mathcal{D})$. To weaken this phenomenon, we weight
$\phi_{k,f}(\mathcal{D})$ higher for smaller values of $k$.

The \emph{normalized discriminability of $\mathcal{D}$ with respect to $f$} which
we define as
\begin{equation}
  \Delta{(\mathcal{D})}_f \coloneqq
  \frac{1}{\lvert X \rvert}\sum_{k=2}^{\lvert X \rvert} \frac{1}{k} \phi_{k,f}(\mathcal{D}).
\end{equation}
The \emph{normalized intrinsic dimensionality of $\mathcal{D}$ with respect to
  $f$} is then given via
\begin{equation}
  \partial{(\mathcal{D})}_f \coloneqq \frac{1}{\Delta{(\mathcal{D})}_f^{2}}.
\end{equation}
The higher this value is for a given feature, the more it contributes to the
intrinsic dimension and as such diminishes the possibility of distinguishing the
data points.

~\cite{Stubbemann2023SelectingFB} provides an algorithm for calculating the
normalized intrinsic dimensionality directly.

\subsubsection{Approximation of Discriminability}
\label{sec:dimensionality:approximation-discriminability}
Unfortunately, an explicit calculation of the discriminability is infeasible for
larger data sets as the algorithm scales quadratically with the number of data
points.
We can, however, use a similar approach to the previously referenced method of
approximating the intrinsic dimension with the help of support sequences to
approximate the discriminability as well.

For a feature $f \in F$ and a support sequence $s$ we call
\begin{equation}
  \Delta(\mathcal{D})_{s,f}^{+}\coloneqq \frac{1}{\lvert X \rvert}
  \left(\sum_{i=1}^l \frac{1}{s_i}\phi_{s_i, f}(\mathcal{D}) +
    \sum_{i=1}^{l-1}\sum_{s_i<j<s_{i+1}} \frac{1}{j}\phi_{s_{i+1}, f}(\mathcal{D})\right)
\end{equation}
the \emph{upper normalized discriminability with respect to $f$ and $s$} and
\begin{equation}
  \Delta(\mathcal{D})_{s,f}^{-}\coloneqq \frac{1}{\lvert X \rvert}
  \left(\sum_{i=1}^l \frac{1}{s_i}\phi_{s_i, f}(\mathcal{D}) +
    \sum_{i=1}^{l-1}\sum_{s_i<j<s_{i+1}} \frac{1}{j}\phi_{s_{i}, f}(\mathcal{D})\right)
\end{equation}
the \emph{lower normalized discriminability with respect to $f$ and $s$}.

We define the \emph{upper/lower normalized intrinsic dimensionality with respect
  to $f$ and $s$} via $\partial (\mathcal{D})_{s,f}^{+}
\coloneqq\frac{1}{\left(\Delta(\mathcal{D})_{s,g}^{-}\right)^{2}}$ and
$\partial(\mathcal{D})_{s,f}^{-}
\coloneqq\frac{1}{\left(\Delta(\mathcal{D})_{s,f}^{+}\right)^{2}}$.
Equipped with these we then can assign each feature their \emph{approximated
  normalized intrinsic dimensionality with respect to $f$ and $s$}:
\begin{equation}
  \partial({\mathcal{D}})_{f}\simeq \partial({\mathcal{D}})_{s,f} \coloneqq
  \frac{\partial(\mathcal{D})_{s,f}^{+} + \partial(\mathcal{D})_{s,f}^{-}}{2}.
\end{equation}

~\cite{Stubbemann2023SelectingFB} provides an algorithm for calculating this
approximation of the normalized intrinsic dimensionality.

\subsection{Experimental Execution and Impact on Intrinsic Dimension}
As we want to demonstrate the effect of intrinsic dimension of the different
data sets on the methods of the reproduced papers we discard features with the
highest (approximated) normalized intrinsic dimensionality.

For this we first extract the logic for loading and preprocessing from every
paper source code and use a concatenation of samples from both train and test
data in the cases where it could not be avoided.
Crucially we do not give the machine learning method more access to
the test data than in the original implementation.

Contemporary machine learning data sets are usually comprised of a matrix.
For graph data, this usually refers to the data of the nodes $X$.
In addition, the connectivity information, given by the adjacency matrix $A$,
and any edge features, are also considered.
However, aggregating this information into a feature matrix by neighborhood
aggregation of the form $A^kX$ (for $k$ a small positive integer) does not
change the qualitative insights provided by the intrinsic dimension, as shown by
previous work~\citep{Stubbemann2023SelectingFB}.
Because additionally many methods themselves perform forms of aggregation, we
have refrained from taking neighborhoods into account.
Therefore we use only the matrix of node features of shape $n\times d$ where $n$
indicates the number of samples and $d$ the number of attributes per sample.
For each data set in our investigation we use the following representation as a
geometric data set as introduced in~\cref{def:geo-data-set}.
The set $X$ is comprised of the $n$ samples $x_i$ where each sample consists of
the attributes $x_i=(x_{i1},\dots, x_{id})$.
We chose the set of component selectors $f_j(x)=x_{j}$ as the set of feature
functions $F$.
Together with the counting measure $\nu(A)=|A|/n$ for a subset
$A\subseteq X$ we complete our special instance of the
geometric data set $\mathcal{D}$.

The sizes of all used node feature matrices can be seen
in~\cref{tab:data set_sizes}.
\begin{table}[H]
  \footnotesize
  \caption{Sizes for all data sets and the research works they appear in, in the
  scope of this work.}
  \label{tab:data set_sizes}
  \begin{center}
    \begin{tabular}{|l|l|l|l|l|}
      \hline
      \textbf{Data Set Name} & \textbf{Nodes} & \textbf{Edges} & \textbf{Features} & \textbf{Paper Names}
      \DTLforeach{data_set_sizes_grouped}
      {\cola=data_set_name, \colb=nodes, \colc=edges, \cold=features, \cole=paper_names}
      {\cr \hline \cola & \colb & \colc & \cold & \cole}
      \cr \hline
    \end{tabular}
  \end{center}
\end{table}

\paragraph{Data set preparation}
For each research paper, we initially extract the essential components for
loading and preprocessing the data sets from the source code supplied by the
authors.
We use these to obtain the node feature matrices of the data sets used.
In instances where it is unavoidable, we resort to concatenating the node
feature matrices from both the training and test data.
An important aspect to note is that we strictly ensure the machine learning
method does not have more access to the test data than what was granted in the
original implementation.

Our rationale for applying feature selection after preprocessing is as follows:
With our approach we want to investigate how methods are influenced by the data
on which they are applied, e.g.\ how the model ``sees'' the data.
Some forms of preprocessing change the empirical data distribution and
preprocessing usually does not contain any learnable parameters.
Additionally the model in question has almost never explicit information about
the applied preprocessing steps.
Thereby we do not consider the preprocessing steps as part of the model.
This makes it easier to disentangle the influence, otherwise we would also
include the change of the preprocessing by the feature selection in the
resulting observations and discussions.

\paragraph{Feature selection}
We used the algorithm for direct calculation of the discriminability
(\cite{Stubbemann2023SelectingFB}, Algorithm 1) for data sets with less than
$10^5$ samples.
For larger data sets, we employed the approximating version
(\cite{Stubbemann2023SelectingFB}, Algorithm 2).
In those cases we first choose a geometric sequence $\hat{s}=(s_1,\dots s_l)$
of length $l=10,000$ with $s_1 = \lvert X \rvert$ and $s_l=2$ and use the
support sequence (\cref{sec:dimensionality:approximation-discriminability}) $s$
which results from $s'=(\lfloor \lvert X \rvert +2 - s_1\rfloor, \dots, \lfloor
\lvert X\rvert + 2 - s_l \rfloor)$ via discarding duplicated elements.
We then discarded for every factor $\alpha\in\{0.1, 0.2, \dots, 0.9\}$ the
corresponding fraction of the features with highest (approximated) normalized
intrinsic dimensionality from all data points.
After the selection we run the machine learning algorithms of the corresponding
papers on the feature reduced data sets with the same (hyper-) parameters
configuration as the original.
For evaluation we collected the same scores as the original works (accuracy or
f1 scores) over repeated training runs with ten different seeds.%
\footnote{An exception was the diffpool enzymes experiment, where only a smaller
  number of runs was feasible given the runtime of the algorithm.}
We do not test other feature selection methods as similar investigations were
already done in~\cite{Stubbemann2023SelectingFB}.

\subsection{Observations}
\label{sec:dimensionality:observations}
We present in this section the computational results and observations for the
experiment.
Here we focus on the details corresponding to the two research works \GCN{} and
\SAGNSLE{}.
Afterwards we will state general observations for the remaining experiments, but refer
the reader to \cref{sec:appendix:more_figures} for accompanying plots.

\begin{figure}
  \centering
  \begin{subfigure}[t]{.49\textwidth}
    \includegraphics[width=\linewidth]{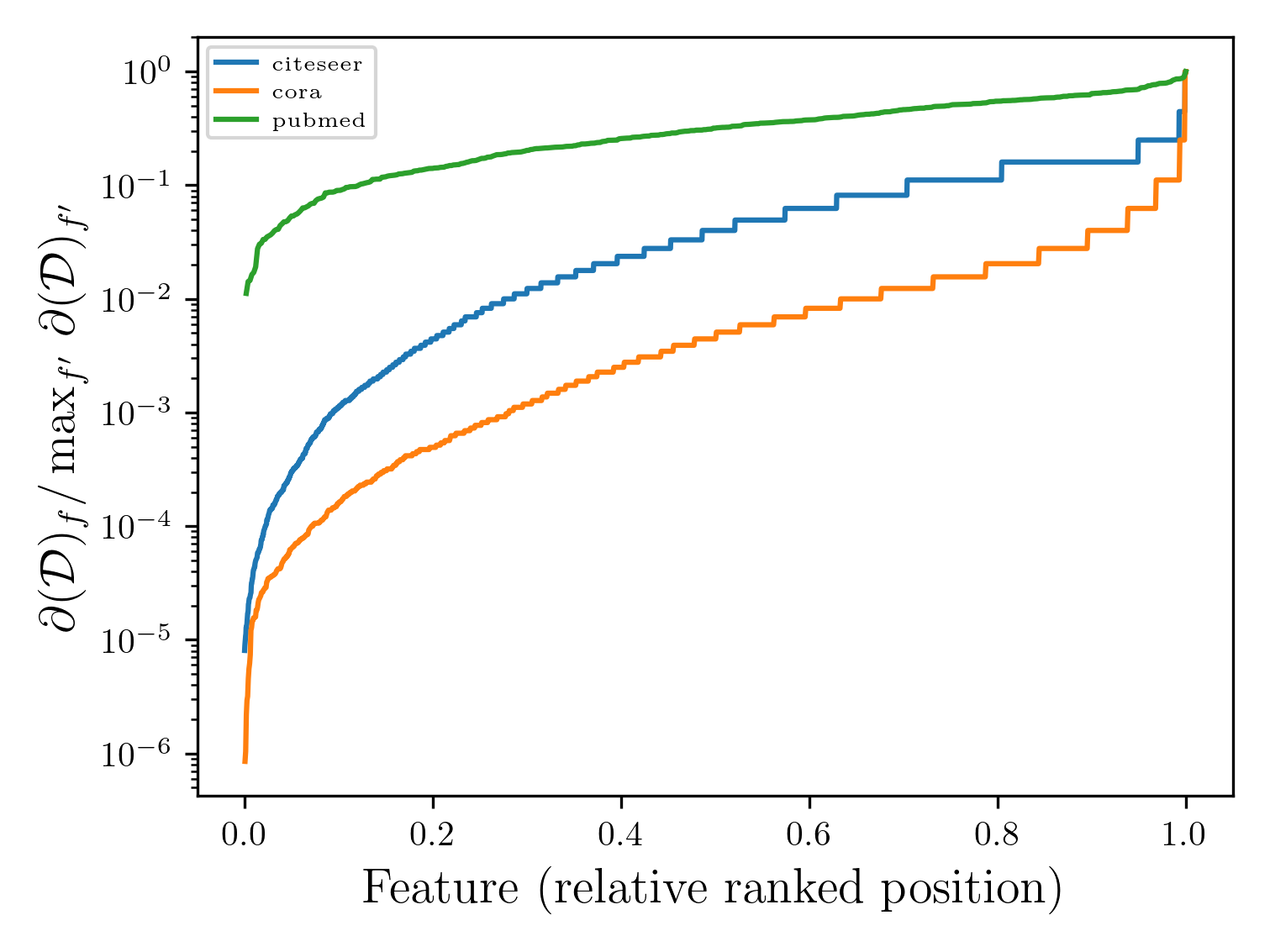}
    \captionsetup{width=.95\linewidth}
    \caption{Normalized intrinsic dimensionality (y-axis) of the \emph{cora},
      \emph{citeseer}, and \emph{pubmed} data sets plotted against relative
      ranked position of features (x-axis).
      For every data set $\mathcal{D}$ the sorting key is defined by normalized
      intrinsic dimensionality divided by $\max_f \partial(\mathcal{D})_f$.
      The values themselves are normalized by the highest value and sorted in
      ascending order.}
    \label{fig:gcn_dimensions}
  \end{subfigure}
  \begin{subfigure}[t]{.49\textwidth}
    \includegraphics[width=\linewidth]{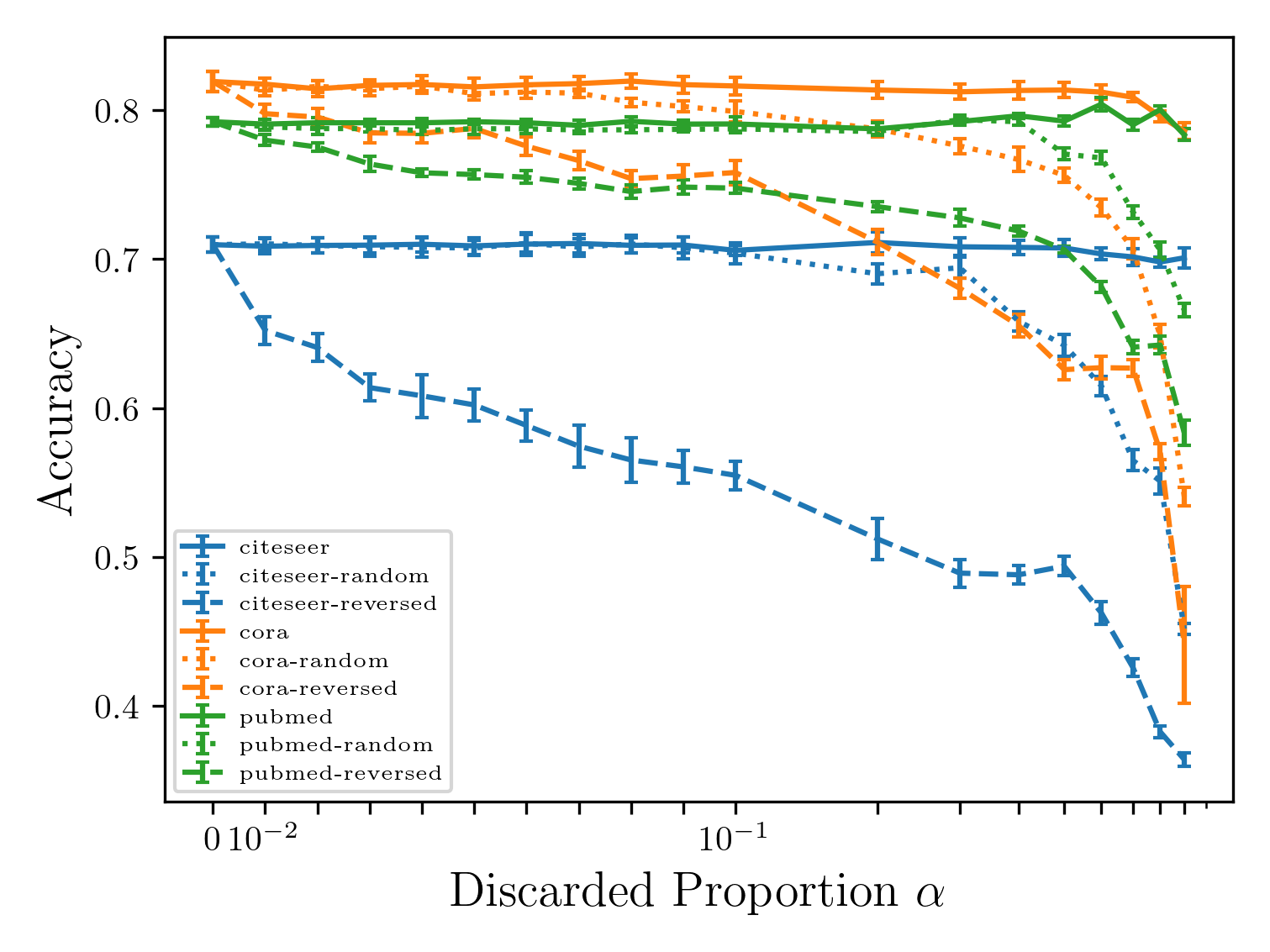}
    \captionsetup{width=.95\linewidth}
    \caption{Accuracy of the resulting model (y-axis) after altering the
      \GCN~data sets based on feature selection.
      We discarded a fraction $\alpha$ of features (x-axis) with the highest
      normalized intrinsic dimensionality from the original data set.
      Curves labeled with \textit{random} or \textit{reversed} used a random or
      reversed selection method respectively.
      Bars indicate standard deviation over ten repetitions with different seeds.}
    \label{fig:gcn_factors}
  \end{subfigure}
  \caption{Influence of Intrinsic Dimension measured through feature selection
    for the \GCN~results.}
  \label{fig:gcn_experiment}
\end{figure}

In \cref{fig:gcn_dimensions} we show the analysis for the intrinsic
dimensionality of the three data sets from \GCN.
Because the data sets are differently sized we need to find a common
representation.
First we order the feature set for each data set using the
normalized intrinsic dimensionality of each feature as the score.
On the x-axis we give the relative position of the sorted feature set, i.e.,
position $\alpha$ indicates that the corresponding feature is at the sorted position
$\alpha\cdot |F|$.
As the measured normalized intrinsic dimensionality can vary widely between the
data sets we decided to normalize it by dividing, for each data set
$\mathcal{D}$, the value $\partial{(\mathcal{D})}_f$ by
$\max_f\partial{(\mathcal{D})}_f$.
The corresponding values are depicted in the y-axis in \cref{fig:gcn_dimensions}.

We observe that all curves increase monotonically in value with respect to the
ranked position of the features.
This is expected as we sort by this value.
However, the slope is solely dependent on the individual contributions of the
features to the intrinsic dimension.
The stair case pattern is not an artifact of the plot but rather results directly
from the data set and its preprocessing.
This indicates that a lot of features have the same
normalized intrinsic dimensionality per step.

We further observe that the \textit{pubmed} data set (green) entails features
with a similar high normalized intrinsic dimension.
Or more general, the higher the line in the plot the more similar are the
values of the individual features of a data set $\mathcal{D}$ compared to the
maximal feature value $\max_f\partial{(\mathcal{D})}_f$.
This allows for comparing the feature behavior of the different data sets.
For example, with respect to this property we observe that the \textit{cora}
data set (orange) has more diverse distributed features compared to the
\textit{pubmed} or \textit{citeseer} data set (blue).

\begin{figure}
  \centering
  \begin{subfigure}[t]{.49\textwidth}
    \includegraphics[width=\linewidth]{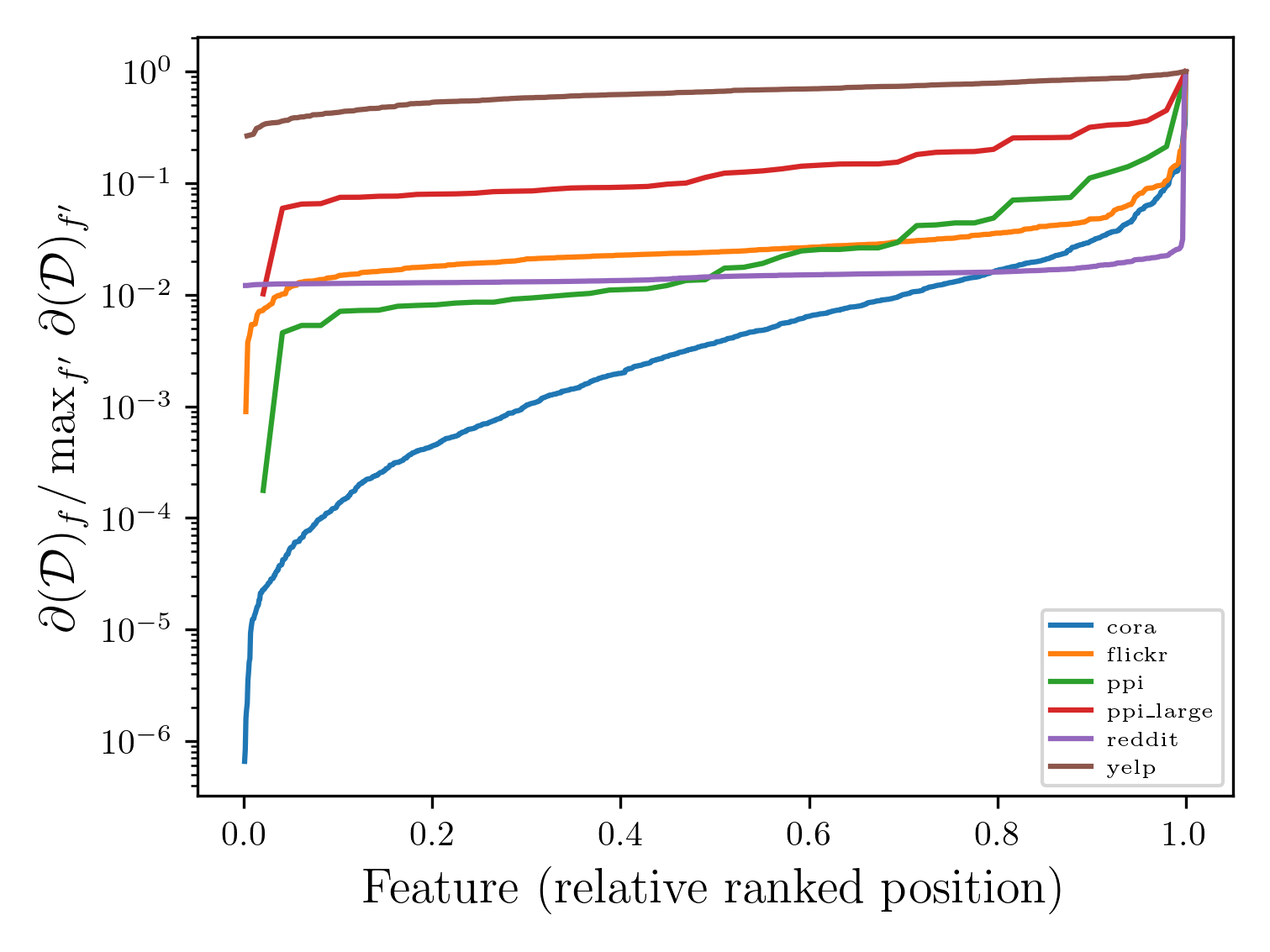}
    \captionsetup{width=.95\linewidth}
    \caption{Normalized intrinsic dimensionality plotted against relative ranked
      position of features.
      See~\cref{fig:gcn_dimensions} for more detailed captions.}
    \label{fig:sagn_with_sle_dimensions}
  \end{subfigure}
  \begin{subfigure}[t]{.49\textwidth}
    \includegraphics[width=\linewidth]{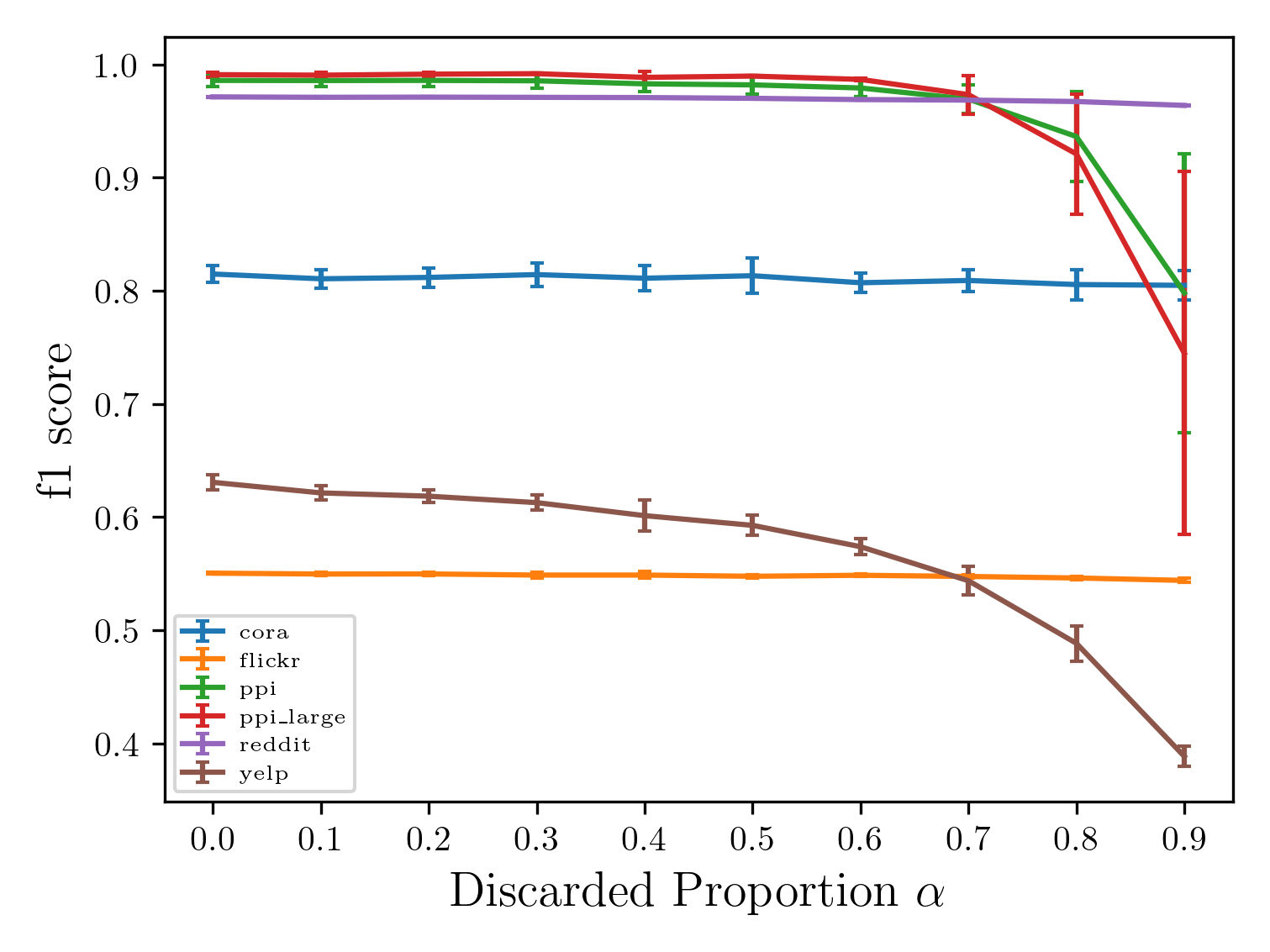}
    \captionsetup{width=.95\linewidth}
    \caption{Performance (measured by f1 score) of resulting model after
      discarding features from the data sets.
      See~\cref{fig:gcn_factors} for more detailed captions.}
    \label{fig:sagn_with_sle_factors}
  \end{subfigure}
  \caption{Influence of Intrinsic Dimension measured through feature selection
    for the \SAGNSLE{} results.}
  \label{fig:sagn_with_sle_experiment}
\end{figure}

\cref{fig:sagn_with_sle_dimensions} demonstrates that the distributions
for the normalized intrinsic dimension of the data sets used for the \SAGNSLE{}
method are of greater variety than those discussed earlier.
We can see that the \textit{cora} data set does not have a prominent stair case
pattern, which can be explained by different preprocessing steps that smooth the
features relative to each other.
One standout distribution is that of the \textit{reddit} data set, which is
shaped like a hockey stick.

As we calculated the (approximated) normalized intrinsic dimensionality on
the data sets, occasionally different normalized rankings for what seems to be
the same data set emerged through different steps of their preprocessing.
A highly visible example can be found in~\cref{sec:appendix:more_figures} with
the \textit{reddit} data set in the \GraphSage~(\cref{fig:graphsage_dimensions}) and
\SGC~(\cref{fig:sgc_dimensions}) experiments.

\paragraph{Accuracy and Intrinsic Dimension}
\cref{fig:gcn_factors} shows the accuracy of the resulting model when applying
the \GCN~machine learning method to the feature reduced data sets.
In this experiment, both the training and test data sets are feature-reduced.
This ensures that the algorithm is trained and tested on the same set of
features.

For each data set we reduced the number of features in steps of 1\%
up until 10\% was reached.
Here the percentage steps are taken with respect to the size of the complete
feature set $F$.
Afterwards we continued with 10\% steps.
Both are indicated on the x-axis in \cref{fig:gcn_factors}.
After training the corresponding model we collected the obtained accuracy,
similar to the original work.
To achieve a meaningful estimate for the model behavior with regard to our
dimensional data set perturbation, we measure the average over ten identical
runs with different seeds.
The resulting standard deviation is shown via error bars in the plot.
Similar plots for the other papers can be found in
\cref{sec:appendix:more_figures}.

We observed that the methods sometimes failed to converge for smaller discarding
values ($<0.01$).
This behavior was very irregular and we did not include these runs and their
corresponding discarding values in the figures.
Our investigation into the causes showed that this was usually due to some
artifacts of the machine learning method, such as early stopping.

Additionally we conduct further experiments with random or reversed feature
selection, where the latter means the discarding of features with the lowest
normalized intrinsic dimensionality first.
For the sake of completeness, by random we refer to the process of randomly
selecting features from $F$.
For all discarded data sets we applied the \GCN~method with ten different seeds.
Due to the expected long run times, these extended experiments (random/reversed,
1\% discard steps) were not conducted for methods other than \GCN.

For all data sets the resulting model performances are relatively stable under
the aforementioned primary discarding method.
Yet when applying the reverse selection method a fast deteriorating performance
can be observed.
This behavior starts already at the smallest discarded proportion and is very
pronounced.

A more intricate detail can be observed for the random discarding method by
combining the information from the two Figures~\ref{fig:gcn_dimensions}
and~\ref{fig:gcn_factors}.
We find that the higher the line in \cref{fig:gcn_dimensions} the later (i.e.,
higher values of $\alpha$) the break off between performance of normal and
random discarding in~\cref{fig:gcn_factors}.

We also see small fluctuations and drops in performance at the highest
discarding values.
This becomes more apparent when directly visualizing the differences to the
proposed discarding method.


This picture changes slightly when looking at the results related to the
\SAGNSLE{} experiments in~\cref{fig:sagn_with_sle_factors}.
For some data sets the same stagnating behavior is evident.
For others, however, there is a marked drop in performance.
Especially for the two \textit{ppi} data sets there is a greater variation in
performance.
Another different behavior can be seen for the \textit{yelp} data set, where the
performance starts to decrease for lower discard factors.

\subsection{Discussion}
\label{sec:dimensionality:discussion}
We now want to contextualize the observations and results.
The~\cref{fig:gcn_dimensions,fig:sagn_with_sle_dimensions}
show the distributions of \ac{NID} and we observe that distinct values arise for different data
sets.
We may note that the figures show relative and not absolute \ac{NID} and therefore
their respective values should not be compared.
Moreover, even without this relative scaling does the mathematical modeling of
the intrinsic dimension not allow a direct comparison.

For our discussion we compare the different values of \ac{NID} to the performance of
the corresponding models, as shown
in~\cref{fig:gcn_factors,fig:sagn_with_sle_factors}.
From this we can infer the following link.
consider the difference between the lowest and the highest value of the \ac{NID}
for a given data set.
We find that this difference decreases in the following order: \textit{pubmed,
  citeseer, cora}.
If we look to the corresponding model performance in~\cref{fig:gcn_factors}, we
observe that the performance of the random feature selection divergences for
different proportions $\alpha$.
Interestingly this happens in the same order as before, albeit at different
levels of accuracy.
The difference between the lowest and highest values of \ac{NID}
indicates that the individual contributions to the \ac{ID} by the corresponding
features is more evenly distributed.
For example, in the case of a horizontal line every feature contributes
equally to the \ac{ID}.
Conversely, a \scalebox{1.9}{$\lrcorner$}-shape, as observed
in~\cref{fig:sagn_with_sle_dimensions} for the \emph{reddit} data set, indicates
that a small number of features is responsible for almost all of the \ac{ID}
value.
Based on these deductions, we propose the following explanation.
In a certain sense, features with a low \ac{NID} can be used by machine learning
methods to distinguish more data points.
These features may allow a learning procedure to have a more stable convergence,
a shorter runtime, and a higher final performance.
In our experimental study, we focus on the interplay between the \ac{ID} and the
achieved model performance.
At this point we may note that all presented experiments are based on the same
principal optimization task of \ac{SGD}.

The observations concerning the shape of distribution of \ac{NID} described
above may permit to draw a connection to the simplicity bias in neural
networks~\citep{Arpit2017ACL,Shah2020ThePO,Valleperez2019DeepLG},
the tendency of \ac{SGD} to find simple models.
Although the need for further investigations arises, the hypothesized link
would then be in line with the assumption that \ac{SGD} weights features that
carry more information higher.

In our random experiment we uniformly discard from the set of features.
In every step one may lose low and high dimensional features following the
distributions as shown
in~\cref{fig:gcn_dimensions,fig:sagn_with_sle_dimensions}.
This means for a certain discarding proportion $\alpha$ there are almost no
features with low \ac{NID} available.
Until those disappear, \ac{SGD} has the possibility to use them for obtaining
the objective.
But when those ``good'' features are not included anymore the situation changes
and the performance deteriorates rapidly.
On the other hand if the features are discarded in order of decreasing \ac{NID}
then the inevitable deterioration of model performance can be postponed for
quite a bit.
Conversely when discarding features in a reverse order, e.g.\ ascending
\ac{NID}, the performance drops rapidly as the model has no access to those
``good'' features from the beginning.

However, we can not exclude the effect of artifacts of the method.
Especially in the reverse case, where for example a non-convergent behavior in
the beginning triggers an early stopping condition that lead to aborting the
optimization routine.
As we regard the methods as black boxes, we have not investigated these
possibilities further.

Turning to \cref{fig:sagn_with_sle_dimensions} we observe a few data sets that
have a similar distribution of \ac{NID} as those in \cref{fig:gcn_dimensions}.
However, it seems that for some the contribution to the total \ac{NID} are
distributed more evenly among the features.
This is particularly evident for the \textit{yelp} data set.
On the other hand the distribution for the \textit{reddit} data set is far more
extreme, where only a small set of features have extraordinary high contribution
to overall \ac{NID}.
This figure also clearly shows the influence of preprocessing on the \ac{NID}
distribution.
Whereas before in \cref{fig:gcn_dimensions} the line for the \textit{cora} data
set was clearly a step function, it has now become a much smoother slope.
It seems that this has almost no influence on the achieved model performance in
both cases.

Based on the definitions of \ac{ID} and \ac{NID}, it is evident that these functions
should have higher values for a data set than for any of its subsets, as long as
the set of feature functions remains constant.
This is clearly demonstrated in the graphs for both versions of the \textit{ppi}
data set, where one is the super set of the other.
It is worth noting that the final model performances for these data sets have a
high standard deviation, which is much higher than in any other experiments.
A definitive cause could not be determined, but it seems reasonable to
assume that some artifact of the method or some form of mode collapse produced
these high variations.
The detailed figures for the remaining experiments can be found in the appendix.
Now we will discuss the insights that can be obtained through an overarching
analysis.
To achieve this, we will resort to an inter-method comparison since only a few
data sets have been processed by multiple papers.
The aim is to identify similar distributions of \ac{NID} and compare the effects
on model performance.
If both methods behave differently on these (possibly different) data sets, this
could indicate that the \ac{NID} has an impact on the methods.

The \GraphSage{} and \SAGNSLE{} methods both use the \textit{reddit} data set
with the same preprocessing.
The former shows a slight deterioration, while the latter shows almost no change
in model performance.

The \SAGNSLE{} method is applied on the \textit{yelp} data set, while the \GCN{}
and \SGC{} methods are used on the \textit{pubmed} data set.
The \ac{NID} distributions are quite similar, but the performance on the
\textit{yelp} data set continuously worsens with higher $\alpha$, while the
performance on the \textit{pubmed} data set remains stable until the highest
discarding proportions.

Both the \GCN{} and \SGC{} methods use the \textit{citeseer} data set with the
same preprocessing.
The first method shows very stable performance, but the second method shows
minor deterioration.

The \GCN{}, \SGC{} and \SAGNSLE{} methods all use the \textit{cora} data set.
Although the preprocessing is the same in \GCN{} and \SGC{}, there is a clear
difference in performance.

The \SAGNSLE{} and \GraphSage{} method both use the large \textit{ppi} data set
mentioned earlier, but in both cases the performance deteriorates significantly,
starting at different discarding proportions and speeds.

One possible alternative hypothesis for the aforementioned observations is that
the machine learning tasks being addressed can be resolved using only the
information provided by the adjacency matrix of the graph data.
In this case, the reduction of available node features would not affect the
final performance.
Contrary to this, observations from experiments with reversed feature discarding
suggest that that the previous statement may not be accurate, as these
observations showed a far greater deterioration of model performance than the
other way around.

In general this highlights a limitation of the current approach, as the chosen
feature functions do not take into account graph edges.
This indirectly connects to the earlier discussion on the modeling choice of
what to use as the underlying set for the geometric data set.
We decided to use the node features as the base set $X$ and ignored the
edge features or the adjacency matrix.
By using a different modeling the set of feature functions could be extended or
constructed completely different.

These comparisons in their entirety form a strong indication that there is an
influence of the \ac{NID} on the model performance.
Although we have only used the \ac{NID} as an auxiliary tool to measure the
\ac{ID}, it shows that different methods are influenced by the \ac{ID} of the
data set itself.
However, it is difficult to quantify the extent of this dependence on the
concentration phenomenon given the present experiments on these very different
methods.

At this point it is necessary to go into more detail regarding the graphs
for the experiments for the \DiffPool{} model.
The original code accompanying the \DiffPool{} publication uses an one-hot
encoding of the node label as node attributes instead of the available original
node attributes.
The paper does not state this in any way and the results do not seem to be
straightforward reproducible when changing to using the conventional node
attributes.
Nonetheless, we continued with this change to make it compatible with the other
experiments and our method.
Therefore the graph for the reached accuracy starts much lower for zero and low
discarding fractions $\alpha$ as the originally claimed performance would
suggest.

\subsection{Summarizing the Analysis and Limitations}
\begin{figure}
  \centering
  \begin{subfigure}[t]{.49\textwidth}
    \includegraphics[width=\linewidth]{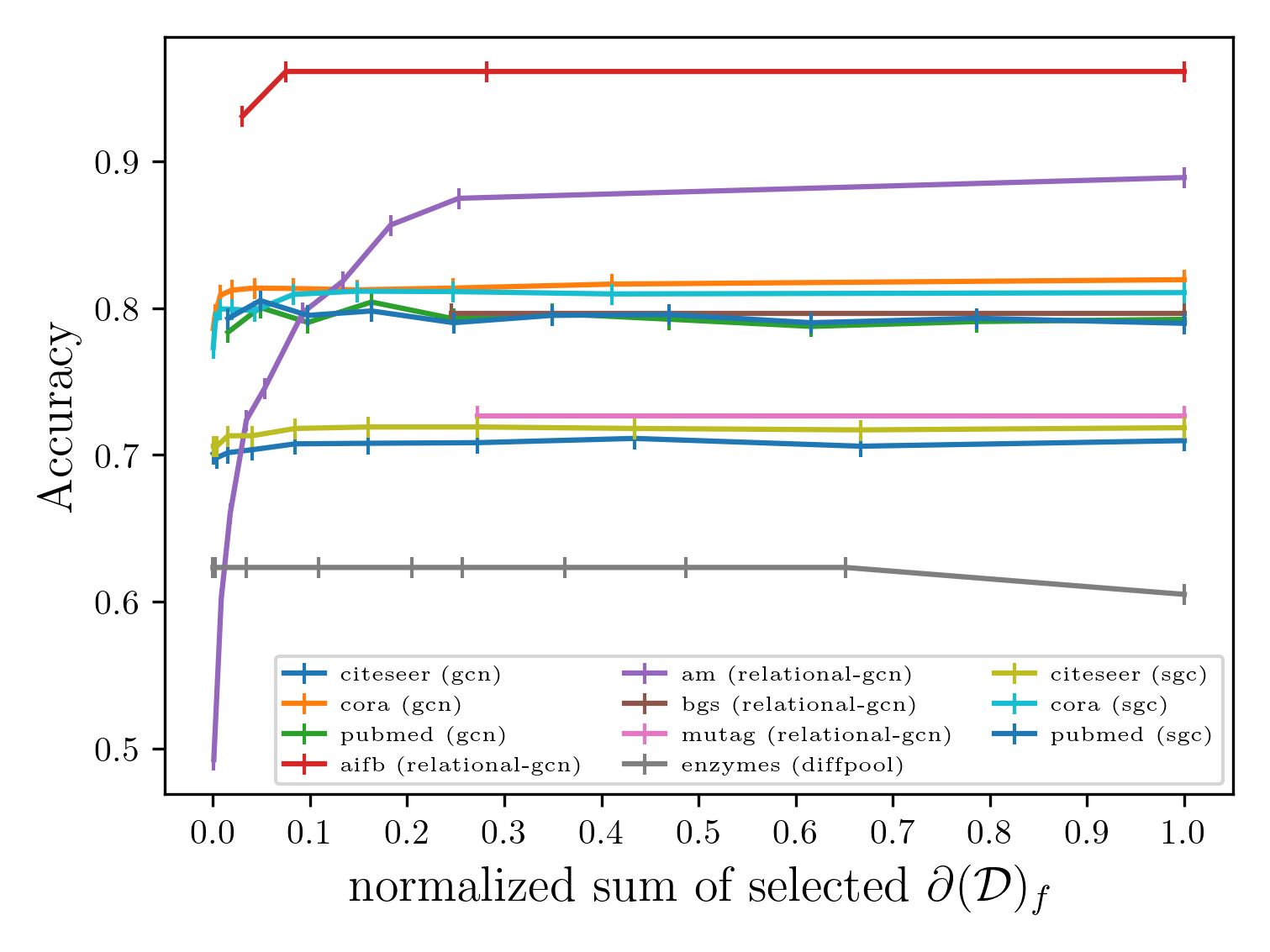}
    \caption{Accuracy of data set and paper combinations}
    \label{fig:reduced_dim_acc}
  \end{subfigure}
  \begin{subfigure}[t]{.49\textwidth}
    \includegraphics[width=\linewidth]{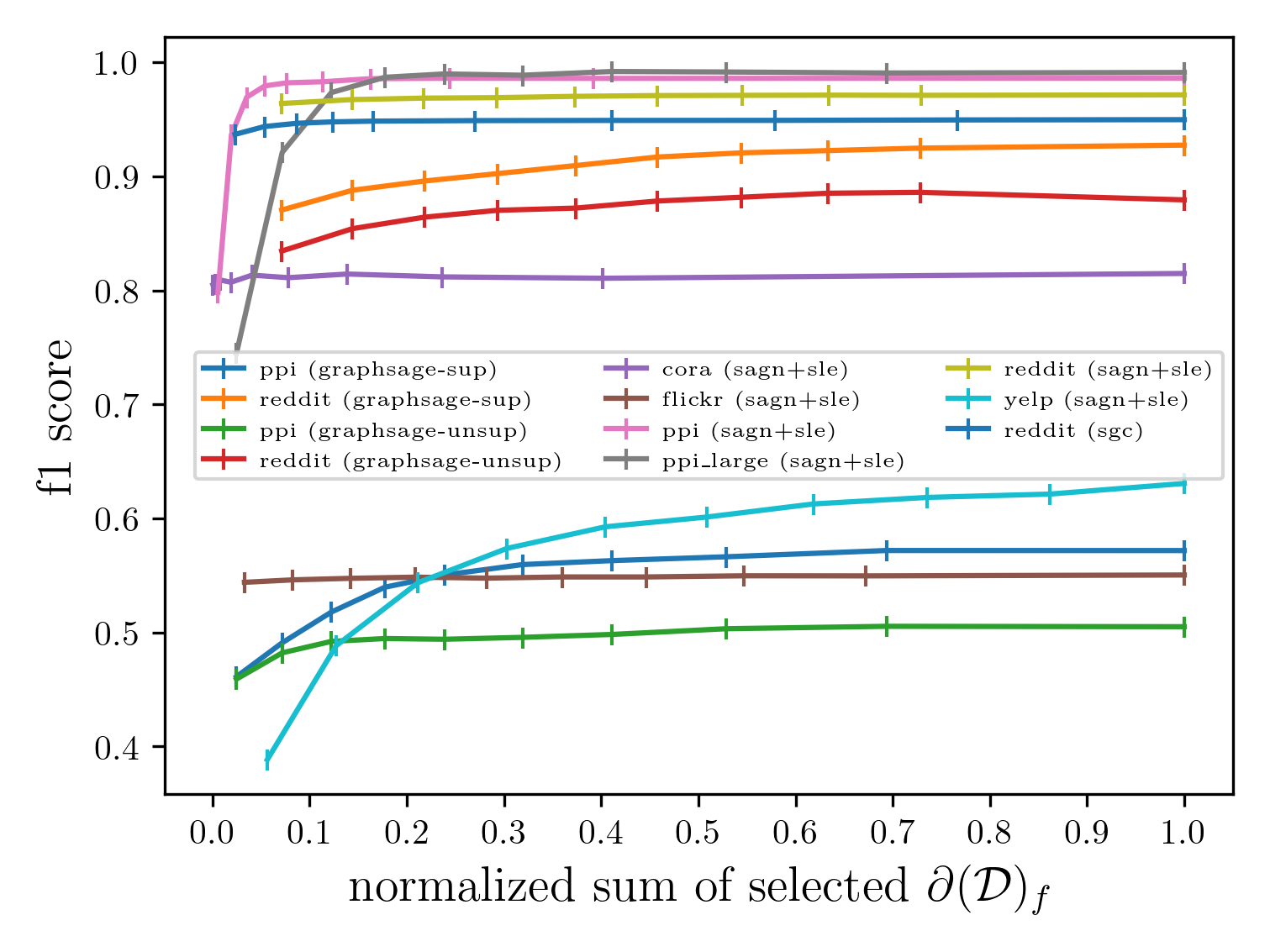}
    \caption{f1 scores of data set and paper combinations}
    \label{fig:reduced_dim_f1}
  \end{subfigure}
  \caption{Overview of evaluation of data set and paper combinations over the
    remaining intrinsic dimensionality. The x-axis is the sum of the
    (approximated) normalized intrinsic dimensionality of the remaining features
    normalized by the total sum for the whole feature set. The y-axis is the
    resulting evaluation score obtained by the method trained on the data set
    with corresponding feature selection.}
  \label{fig:reduced_dim}
\end{figure}

We present an overview over all experiments by combining the information
about the intrinsic dimensionality and the model performance when
discarding features.
For this, we calculate the sum of the (approximated) normalized discriminability
of remaining features after discarding a fraction $\alpha$.
The so obtained value can be normalized by the total sum of (approximated)
normalized discriminability of all features.
We perform this calculation for different values of $\alpha$ and all data sets.
We plot these values against the achieved performance measure
(accuracy or f1 score) for the corresponding configuration.
The results are depicted in \cref{fig:reduced_dim}.

From this we can derive the striking observation that most models can cope with
data sets that are reduced to about $30\%$ of their original dimensionality
without performance loss.
Despite the large differences in the number of samples and features among the
data sets, we cannot observe a significant correlation between these
characteristics and the change in model performance.

Through experiments with reversed feature selection (\cref{fig:gcn_experiment})
it became evident that there is an effect of the intrinsic dimension on the
different learning methods.
More specifically we observed that this effect depends on the particular method,
e.g.~\textit{reddit} data set as discussed in \cref{sec:dimensionality:discussion}.

Our study is limited in various aspects, which we will discuss below.
We observe in our investigation a low frequency of overlapping use of data sets.
This is a result of the selection process and the underlying requirements for
reproducibility.
Hence, comparing different methods on the dimensionally reduced data sets is hard.
However, the data sets \textit{citeseer}, \textit{cora},
\textit{pubmed} and \textit{reddit} have non-trivial support over the
considered papers.
For them we see very similar behavior as described in the previous paragraph
(cf.~\cref{fig:reduced_dim}).

Furthermore our results build upon a rather small set of selected candidates.
This could be tackled with allocating more time for achieving reproducibility
per paper, which would allow for fixing or circumvent reproducibility barriers
by searching for the right combination of technical tricks.
It might also be possible to apply the individual methods from the publications
to the other data sets as well.

The proposed method only indirectly measures the effects of the concentration of
measure phenomenon through the perspective of the geometric intrinsic dimension.
Additionally we can not give extensive overview over complete \ac{ID} influence.
This is due to the fact that it is unclear if the \ac{ML} methods can draw on
more (complicated) feature functions than the considered ones their processing
of the data.
In this case, the current restriction would be a hindrance to measuring the true
impact of the \ac{ID} on the methods.
However, for the used function class we can build upon the guaranteed
computability.
Looking back on the results, it might not be necessary as in even this
restricted scope a influence could be showed.

Furthermore the present work includes no comparison with other approaches of
measuring dimension influence and any feature selection methods based on it.
Although it is not so clear if those methods would measure the same properties
of the \ac{ML} methods given their different underlying theoretical frameworks.

Overall the experiments show that the distribution of dimensionality
contributes to the deterioration of the model performance.
Yet, more experiments are necessary to determine a thorough characterization of
this dependency, especially by using more extensive feature function classes
that capture the geometric intrinsic dimension more thoroughly.

\section{Reproducibility of the Presented Work}
As a publication about reproducibility is is only fair to also consider the own
reproducibility as well.
In general the reproducibility of our own work is limited by the reproducibility
of the used papers.
We rely exclusively on data sets provided by them and our source code build upon
on the one published by them.
For our experiments we do not need to change the (hyper-) parameter of models.
To achieve high degree of reproducibility we provide the individual scripts for
the reproducibility and dimensionality experiments.
This includes explicitly specified environments that were used and the
necessary changes to the original source code.
We tried to follow general
guidelines~\citep{improving_reproducibility,research_code_publishing} and our
own ontology.
In total the experiments produced around 600 models (6 papers, 10 factors, 10
seeds).
It is not feasible to provide weights for all of them, especially if the
original source code did not include sophisticated management of paths where
checkpoints were saved to.
Given the special form of our publication we do not provide the model weights.
The complete resources for source code, logs, and results of our experiments
can be found at \url{https://zenodo.org/doi/10.5281/zenodo.10727907}.

\section{Conclusions}
In this work, we presented an ontology to investigate reproducibility of machine
learning research. This ontology can be used to evaluate reproducibility of
scientific publications in a standardized manner. For this, we assume that
scientific evidence can be generated via theoretical evidence (for example via
theorems and proofs) or via empirical evidence provided by scientific
experiments. In machine learning, reproducibility research mainly focuses on the
extend to which other researchers can re-execute the experiments with the same
results. Thus, our work mainly focuses on empirical evidence for which we
propose concrete attributes for evaluating the level of reproducibility of a
specific work. These attributes can be separated into three categories, namely
\textit{data set}, \textit{software} and \textit{computational result}.

If reproducibility is given, the next step is to identify relevant influential
factors for the experiments outcome. Such a factor which is commonly
investigated is the intrinsic dimension, which measures the influence of the
\cod~on a specific dataset. In the second part of our work,
we investigated to which extent results of empirical experiments depend on the
intrinsic dimension of the datasets used for training. To be more detailed, we
analyzed wether changes on the intrinsic dimensionality of used data sets
coincide with experiment outcomes, measured by model performance.

To give a practical example on the usage of our ontology, we provided detailed
descriptions of how the attributes of our ontology can be used to evaluate the
reproducibility of research on graph neural networks. Furthermore, we
investigated which of the well-known methods are not affected by modifications
of the intrinsic dimensionality of the datasets on which they are trained on.

Our investigation shows that attributes of reproducibility which are part of the
\textit{data set} category of our ontology are equally covered by most works on
graph neural networks. Here, we only found major differences between the
different methods when it comes to the extend to which the datasets are
documented as part of the README. The amount of documentation was one of the key
factors on successful reproduction with a reasonable effort. The documentation
of a specific work is covered by our ontology in the category \textit{software}.
In general, the \textit{software} category allowed for an accurate
differentiation of the different papers with regard to their reproducibility.

Attributes falling under the category \textit{computational result} can be split
into two parts. One part consists of weakly separating attributes which either
present basic reproducibility rules which are fulfilled by all of our methods or
strict requirements that none of the investigated methods satisfy. The other
part consist of attributes which enable the distinction of the different
methods, as these attributes are only covered by a part of the investigated
methods.

The findings presented in the study in the second part of our work provide
compelling evidence that the \ac{ID} has an influence on \ac{ML} algorithms. To
be more detailed, our experiments show that dropping features with high
individual \ac{ID} has a varying impact on model performance. For example, for
the \GCN~method, dropping high-dimensional features does not fundamentally
decrease accuracies. In contrast, dropping these features when learning with
\GraphSage~leads to stronger deterioration of performance. On the other hand,
dropping features with low individual \ac{ID} first leads to stronger
performance drops for all methods. This indicates that current used graph
learning approaches are susceptible to changes of the intrinsic dimensionality
while their robustness towards the discarding of non-discriminative features
(i.e., features with high individual \ac{ID}) strongly varies.

\section{Recommendations}

Based on our findings, we recommend the following points as best practice for
reproducible machine learning research.
\begin{itemize}
\item \textbf{Write one main script that does everything necessary.} That means
  setup/teardown or preparation of the compute environment, downloading and
  preprocessing of data sets, and running of all reported experiments. It might
  be beneficial to chose an existing software package when handling larger data
  or when working on compute clusters.
\item \textbf{Log all relevant information into files.} This includes all
  results, used input parameters and also chosen hyperparameters if a form of
  hyperparameter optimization was done. Often, such information is only printed
  to the output terminal, which hinders reproducibility.
\item \textbf{Store checkpoints of all used models.} If multiple runs with only
  different seeds are desired, the wrapping script/software should make sure
  that no computed checkpoint is overwritten in the next iteration.
\item \textbf{A workflow management systems is helpful for automated aggregation
    and evaluation of outputs and creation of visualizations.} The
  reproducibility of a research paper is strongly enhanced if all steps from the
  experiment to the final paper, including reported results, tables, and
  displayed figures, can be automatically created by a dedicated script which is
  part of the published code. This full procedure can easily be implemented via
  a workflow management system.
\end{itemize}

Furthermore, we give three recommendations on how to account for the concept of
intrinsic dimensionality in the scope of machine learning research.

\begin{itemize}
\item \textbf{Consider the intrinsic dimensionality of the used training data
    sets when comparing machine learning algorithms.} Our experiments indicate
  that well-established graph neural network approaches heavily suffer from
  increasing intrinsic dimensionality of the input data. Hence, when comparing
  their performance, it is crucial to know the \ac{ID} of the used data to
  estimate to which extent performance differences are caused by the \cod.
\item \textbf{Investigate if discarding of high-dimensional features is possible
    for the chosen GNN.} For certain graph neural network models it is possible
  to discard a significant fraction of features with high normalized intrinsic
  dimensionality without a fundamental drop in performance. For example, when
  regarding the \GCN~model, dropping up to 70\% of the total number of features
  is possible without decreasing accuracies, while this is not possible for
  \GraphSage.
\item\textbf{Account for transformations of the \ac{NID}-distributions induced
    via preprocessing.} Preprocessing of features usually incorporates global
  interactions between them (i.e., averages), which changes the \ac{NID}
  distribution in non-trivial ways. One such case was observed for example for
  the \textit{reddit} data set, where both \SAGNSLE~and \SGC~applied different
  preprocessing procedures which resulted in widely different \ac{NID}
  distributions. As mentioned above, this change of the \ac{NID} distributions
  can fundamentally influence model performance.
\end{itemize}

\section{Limitations and Future Work}
Our current study has some limitations which we will address in the future.
First, as discussed in~\cref{sec:survey:discussion}, our ontology is limited to
a fixed granularity. We realized that the varying needed depth of
reproducibility analysis can not be represented by the proposed ontology. Thus,
we will develop a a hierarchy of ontologies with different fine-grains by
further splitting or aggregating current attributes. For example. depending on
the amount of different experiments provided by a paper, it may be reasonable to
reformulate \textbf{R2} to \textbf{R4} into more detailed cases to capture
different error ranges and fractions of not reproducible results.

Second our current study is concentrated on one specific concept of intrinsic
dimensionality. However, as discussed
in~\cref{sec:related_work:intrinsic_dimension_and_feature_selection}, a variety
of different \ac{ID} estimators exist. Hence, future work will investigate which
of the reported observations generalize over different concepts for intrinsic
dimensionality.

Thirdly, our approach only considers intrinsic dimensionality for a specific
feature sets, namely the usual coordinate projections. However, different
machine learning methods may incorporate different aspects of the data. Thus,
future work will investigate how these different aspects can be formalized as
feature functions. This will lead to an \ac{ID} not on for models instead of dat
sets. Here, the crucial problem will be the identification of a finite and
computational feasible feature set which captures the model behavior.

\section*{Acknowledgment}
The authors thank the State of Hesse, Germany for funding this work as part of
the LOEWE Exploration project ``Dimension Curse Detector'' under grant
LOEWE/5/A002/519/06.00.003(0007)/E19.


\bibliography{bibliography2}
\bibliographystyle{tmlr}

\newpage
\clearpage
\appendix
\section{Considered Publications}
In our survey, we began by collecting a list of publications to consider
(\cref{sec:candidate_selection}).
We utilized the \emph{Semantic Scholar API}, employing the search term ``graph
neural network'' to obtain a set of results.
These results were then processed with a dedicated script to calculate the scoring
metric $\frac{\text{number of citations}}{2023-\text{year of publication}}$ for
each publication.
Out of the vast array of publications, we selected the top 100 papers based on
these scores.

The process of reproducing the list proved to be a challenge, because of a
significant degree of variability due to the non-deterministic nature of the
Semantic Scholar API.\
This variability was particularly noticeable with regard to page ordering and
the contents of the first 10000 entries.
Over time we started reproducibility attempts of publications that are now no
longer part of the list.

In our filtering process, we manually excluded entries that were surveys, coding
frameworks, or lacked a clear connection to graph neural networks.
We also ignored works that only applied Graph Convolutional Network (GCN)
methods to other field of science or used time-dependent or spatial data,
especially in the field of chemistry.
The reason for this was that the feature selection approach used later was not
directly applicable to such work, or at best uninformative.
Additionally, publications applying methods to very specific data sets were not
included in our list.

Some well-known papers, such as \emph{GraphSAINT}, were not included because
they did not appear under the search term used.
This absence also explains why \emph{SAGN+SLE}, \emph{SGC}, and \emph{GraphSAGE}
do not appear in the list.
To better cover the field of graph neural network research, additional search
terms like ``graph convolutional network'' would be necessary.
The subsequent changes in criteria led to a high rate of skipped
publications in the full list, which was generated end of May 2023.

\newcommand{\statusincluded}{i}
\newcommand{\statusexcludedexperiment}{e}
\newcommand{\statusexcludedmethod}{m}
\newcommand{\statusexcludeddataset}{d}
\newcommand{\statusexcludedbug}{b}
\newcommand{\statusexcludedhardware}{h}
\newcommand{\statusskipped}{s}


\begin{small}
\begin{longtable}{| p{.80\textwidth} | p{.07\textwidth} | p{.05\textwidth} |}
  \captionsetup{justification=centering}
  \caption{All considered publications with indication on survey status.
  The indicators are as follows: \\
  \textbf{\statusincluded}~- included in survey,
  \textbf{\statusexcludedexperiment}~- excluded because experiment is not
  available, \\
  \textbf{\statusexcludeddataset}~- excluded because method is only applied on
  unusual or specially build graph data sets, \\
  \textbf{\statusskipped}~- skipped because of time constraints.}
\label{tab:survey:considered} \\
\hline
Publication & Score & Status \\
\hline
Semi-Supervised Classification with GCNs~\citep{Kipf2016SemiSupervisedCW} & 2832.57 & \statusincluded{} \\
\hline
Graph Attention Networks~\citep{Velickovic2017GraphAN} & 1869.83 & \statusexcludedexperiment{} \\
\hline
How Powerful are GNNs?~\citep{Xu2018HowPA} & 913.2 & \statusskipped{} \\
\hline
Modeling Relational Data with GCNs~\citep{Schlichtkrull2017ModelingRD} & 540.0 & \statusincluded{} \\
\hline
LightGCN: Simplifying and Powering Graph Convolution Network for Recommendation~\citep{He2020LightGCNSA} & 531.0 & \statusskipped{} \\
\hline
Neural Graph Collaborative Filtering~\citep{Wang2019NeuralGC} & 417.0 & \statusexcludeddataset{} \\
\hline
Heterogeneous Graph Attention Network~\citep{Wang2019HeterogeneousGA} & 355.5 & \statusskipped{} \\
\hline
Hierarchical Graph Representation Learning with Differentiable Pooling~\citep{Ying2018HierarchicalGR} & 322.4 & \statusincluded{} \\
\hline
Graph Contrastive Learning with Augmentations~\citep{You2020GraphCL} & 303.67 & \statusskipped{} \\
\hline
KGAT: Knowledge Graph Attention Network for Recommendation~\citep{Wang2019KGATKG} & 281.5 & \statusskipped{} \\
\hline
GCNs for Text Classification~\citep{Yao2018GraphCN} & 261.4 & \statusexcludeddataset{} \\
\hline
Link Prediction Based on GNNs~\citep{Zhang2018LinkPB} & 248.4 & \statusskipped{} \\
\hline
GCNs for Hyperspectral Image Classification~\citep{Hong2020GraphCN} & 234.0 & \statusexcludeddataset{} \\
\hline
DeepGCNs: Can GCNs Go As Deep As CNNs?~\citep{Li2019DeepGCNsCG} & 230.75 & \statusskipped{} \\
\hline
E(n) Equivariant GNNs~\citep{Satorras2021EnEG} & 220.5 & \statusskipped{} \\
\hline
Weisfeiler and Leman Go Neural: Higher-order GNNs~\citep{Morris2018WeisfeilerAL} & 211.0 & \statusskipped{} \\
\hline
Predict then Propagate: GNNs meet Personalized PageRank~\citep{Klicpera2018PredictTP} & 210.6 & \statusskipped{} \\
\hline
Heterogeneous Graph Transformer~\citep{Hu2020HeterogeneousGT} & 209.0 & \statusskipped{} \\
\hline
Heterogeneous GNN~\citep{Zhang2019HeterogeneousGN} & 203.0 & \statusskipped{} \\
\hline
Geom-GCN: Geometric GCNs~\citep{Pei2020GeomGCNGG} & 196.67 & \statusskipped{} \\
\hline
Session-based Recommendation with GNNs~\citep{Wu2018SessionbasedRW} & 193.0 & \statusexcludeddataset{} \\
\hline
Self-Attention Graph Pooling~\citep{Lee2019SelfAttentionGP} & 180.5 & \statusskipped{} \\
\hline
GCC: Graph Contrastive Coding for GNN Pre-Training~\citep{Qiu2020GCCGC} & 178.0 & \statusskipped{} \\
\hline
Few-Shot Learning with GNNs~\citep{Satorras2017FewShotLW} & 169.33 & \statusskipped{} \\
\hline
Dynamic Edge-Conditioned Filters in CNNs on Graphs~\citep{Simonovsky2017DynamicEF} & 168.67 & \statusexcludedexperiment{} \\
\hline
How to Find Your Friendly Neighborhood: Graph Attention Design with Self-Supervision~\citep{Kim2022HowTF} & 159.0 & \statusskipped{} \\
\hline
Beyond Homophily in GNNs: Current Limitations and Effective Designs~\citep{Zhu2020BeyondHI} & 158.33 & \statusskipped{} \\
\hline
GNN-Based Anomaly Detection in Multivariate Time Series~\citep{Deng2021GraphNN} & 150.0 & \statusexcludeddataset{} \\
\hline
DKN: Deep Knowledge-Aware Network for News Recommendation~\citep{Wang2018DKNDK} & 149.8 & \statusexcludeddataset{} \\
\hline
MixHop: Higher-Order Graph Convolutional Architectures via Sparsified Neighborhood Mixing~\citep{Abu-El-Haija2019MixHopHG} & 142.0 & \statusskipped{} \\
\hline
MAGNN: Metapath Aggregated GNN for Heterogeneous Graph Embedding~\citep{Fu2020MAGNNMA} & 137.67 & \statusskipped{} \\
\hline
Principal Neighbourhood Aggregation for Graph Nets~\citep{Corso2020PrincipalNA} & 132.67 & \statusskipped{} \\
\hline
Hypergraph Neural Networks~\citep{Feng2018HypergraphNN} & 128.6 & \statusskipped{} \\
\hline
Graph Transformer Networks~\citep{Yun2019GraphTN} & 128.25 & \statusskipped{} \\
\hline
Beyond Low-frequency Information in GCNs~\citep{Bo2021BeyondLI} & 125.0 & \statusskipped{} \\
\hline
Encoding Sentences with GCNs for Semantic Role Labeling~\citep{Marcheggiani2017EncodingSW} & 123.17 & \statusexcludeddataset{} \\
\hline
Neural Motifs: Scene Graph Parsing with Global Context~\citep{Zellers2017NeuralMS} & 122.67 & \statusexcludeddataset{} \\
\hline
Graph Convolution over Pruned Dependency Trees Improves Relation Extraction~\citep{Zhang2018GraphCO} & 122.4 & \statusexcludeddataset{} \\
\hline
Graph Structure Learning for Robust GNNs~\citep{Jin2020GraphSL} & 121.33 & \statusskipped{} \\
\hline
Towards Deeper GNNs~\citep{Liu2020TowardsDG} & 118.67 & \statusskipped{} \\
\hline

\end{longtable}
\end{small}

\section{Reproducibility Context}
\label{sec:appendix:rep_context}
The data collected during the successful reproducibility attempts
in~\cref{sec:survey} can be summarized in a \emph{Formal
  Context}~\citep{Ganter1997FormalCA}, where the papers make up the object set
and the analyzed features of reproducibility the attribute set.
A cross for paper $p$ and feature $f$ means that this feature was observed for
the paper.
Based on the definitions of the features, this indicates a negative aspect of
reproducibility occurring in the paper.

\renewcommand{\longtextorientation}{270}
\explanationtrue
\begin{table}[H]
  \begin{center}
    \captionsetup{justification=centering}
    \caption{Formal Context derived from Reproducibility Survey with adjoining
      error ontology.\\
      The table is rotated sideways, e.g. Papers are objects and reproducibility
      features are attributes.}
    \label{tab:context:full}
    \resizebox{11cm}{!}{
      \begin{sideways}
        \contexttable
      \end{sideways}
    }
  \end{center}
\end{table}

\section{Presentation of other Experiments}
\label{sec:appendix:more_figures}
Here we want to include figures presenting the results from the influence
experiments not yet presented in detail.
The diagrams are structured the same way as the ones presented in
\cref{sec:dimensionality:observations}
For each experiment we first show the normalized distribution of normalized
intrinsic dimensionality for the used data sets (after preprocessing).
For data sets with more than $10^5$ samples, an algorithm for calculating the
approximated \ac{NID} is used.
Additionally a second figure presents the accuracy/f1 scores obtained when
training on the feature reduced data sets.
For a more detailed explanation see description accompanying
\cref{fig:gcn_experiment}.

\begin{figure}[H]
  \centering
  \begin{subfigure}{.49\textwidth}
    \includegraphics[width=\linewidth]{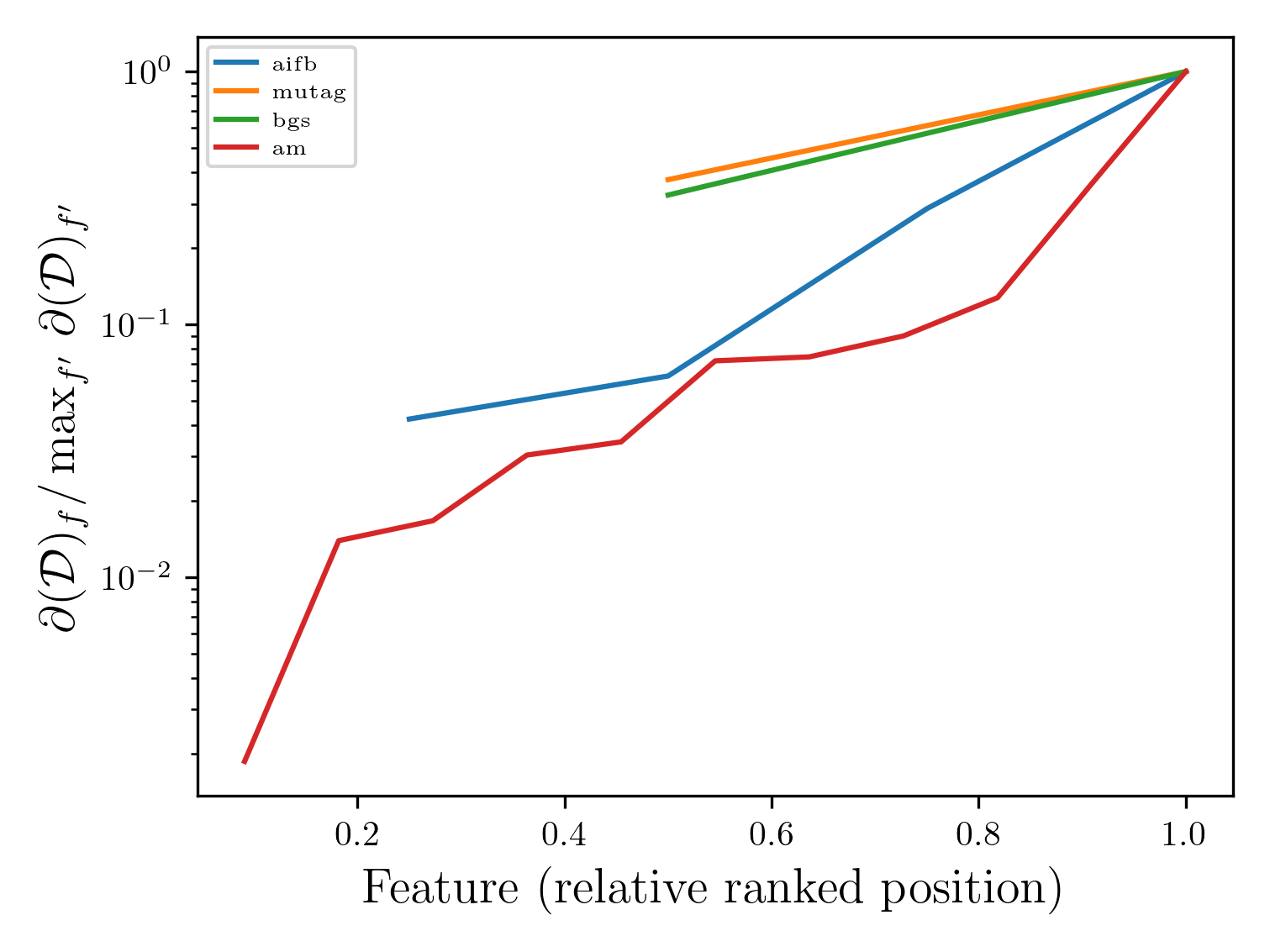}
    \caption{Normalized intrinsic dimensionality.}
    \label{fig:rgcn_dimensions}
  \end{subfigure}
  \begin{subfigure}{.49\textwidth}
    \includegraphics[width=\linewidth]{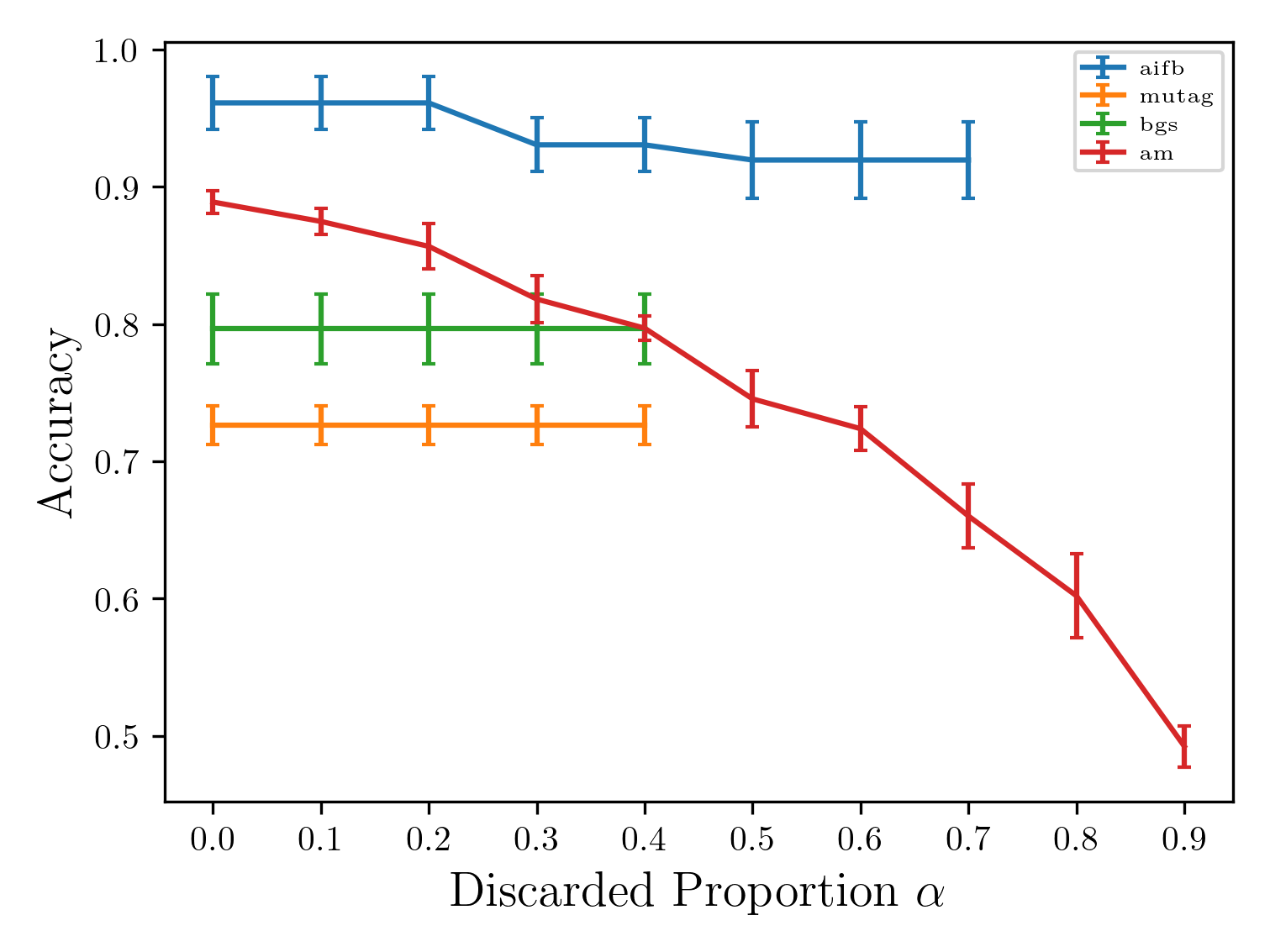}
    \caption{Performance of resulting model.}
    \label{fig:rgcn_factors}
  \end{subfigure}
  \caption{Experiment based on \RGCN.}
  \label{fig:rgcn_experiment}
\end{figure}

Some data sets have so few features that the steps of the discarding process
are smaller than one feature.
This leads to fewer data points, which in turn give the impression of only
partially complete graphs for visualizations of normalized distribution of
the \ac{NID} or accuracy for given discarding proportions as the necessary
granularity can not be achieved (see~\cref{fig:rgcn_experiment}).

\setlength{\belowcaptionskip}{-10pt}

\begin{figure}[H]
  \centering
  \begin{subfigure}{.49\textwidth}
    \includegraphics[width=\linewidth]{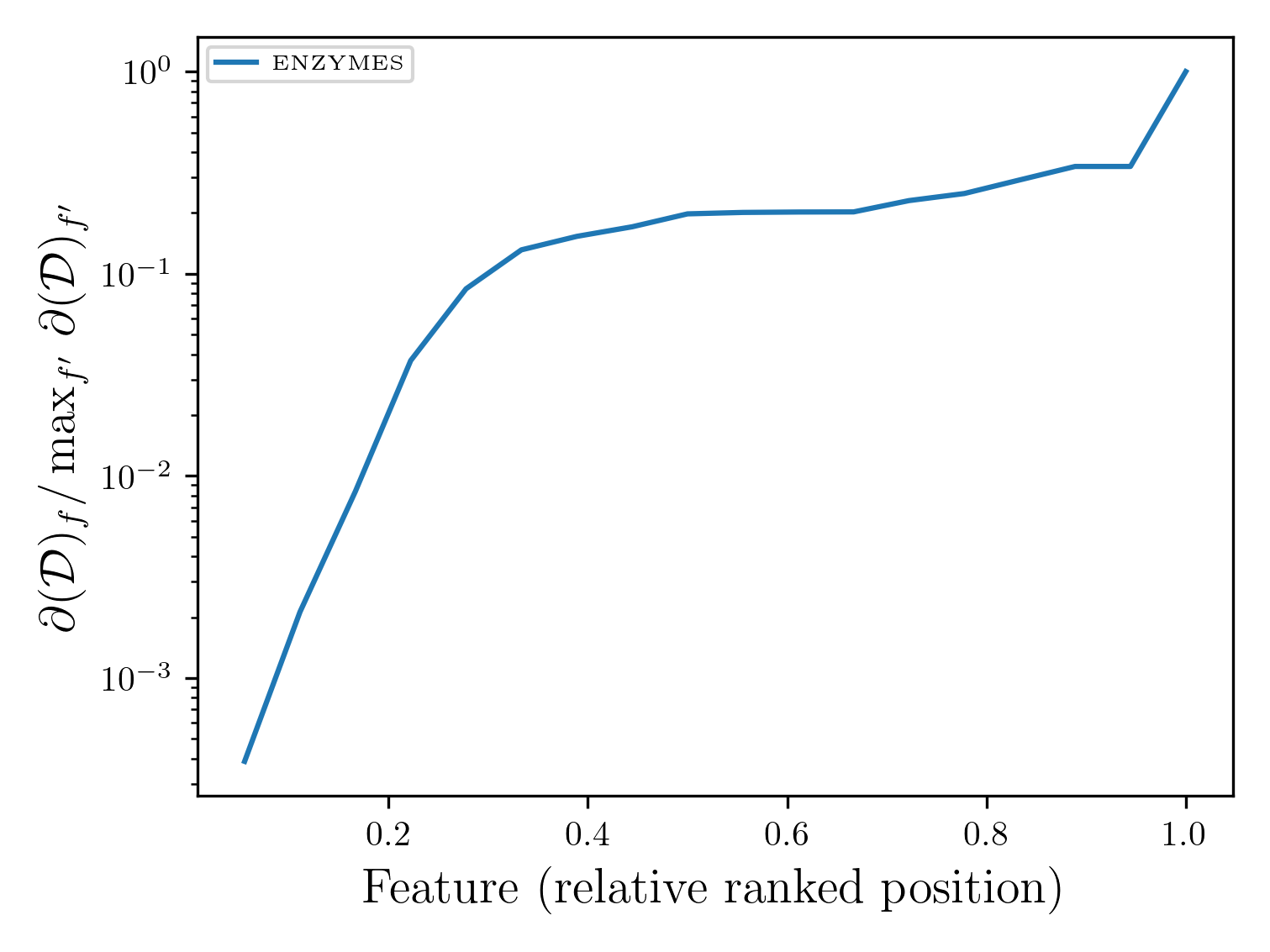}
    \caption{Normalized intrinsic dimensionality.}
    \label{fig:diffpool_dimensions}
  \end{subfigure}
  \begin{subfigure}{.49\textwidth}
    \includegraphics[width=\linewidth]{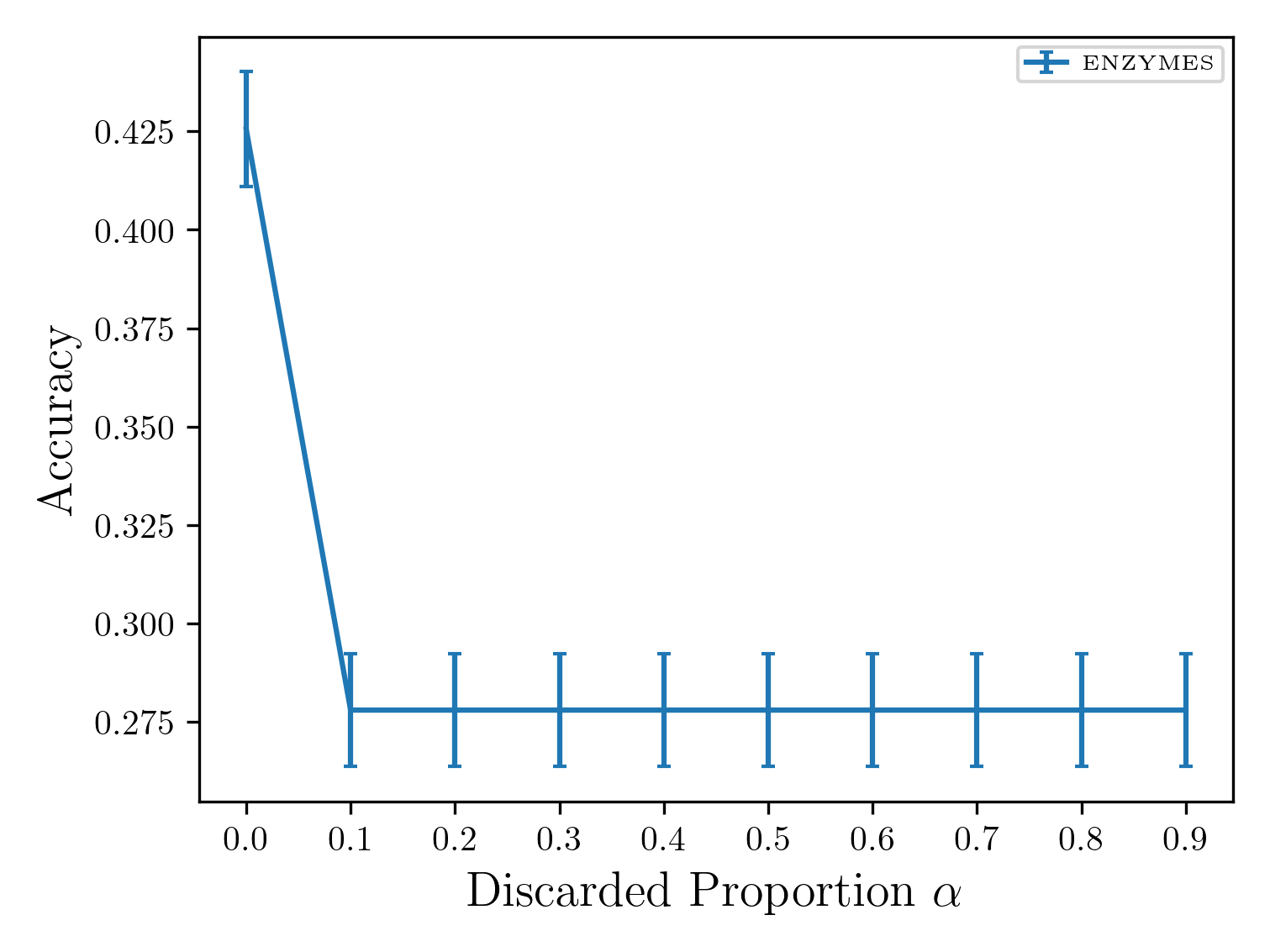}
    \caption{Performance of resulting model.}
    \label{fig:diffpool_factors}
  \end{subfigure}
  \caption{Experiment based on \DiffPool.}
  \label{fig:diffpool_experiment}
\end{figure}

The \DiffPool~publication used only two data sets, of which only one, namely the
\emph{enzymes} data set, has features for the graph nodes.
Therefore we were not able to apply the described method to the other data set.
Furthermore we encountered another problem during the \DiffPool~experiments
(see~\cref{fig:diffpool_experiment}).
It is strongly implied in the paper that the method uses the node features in its
computations.
However, a close examination of the source code reveals that in the default
configuration, the node features are built from the classification targets of
the associated graph.
By changing the corresponding parameter in the training script to a different
argument, which we decided on the basis of which preprocessing modifications it
induced, no improvements could be observed.
On the contrary, the overall performance of the method got worse.
Nevertheless, we present the results obtained, which again show, that the
\DiffPool~method does not use the node features in a comprehensible way.

\begin{figure}[H]
  \centering
  \begin{subfigure}{.49\textwidth}
    \includegraphics[width=\linewidth]{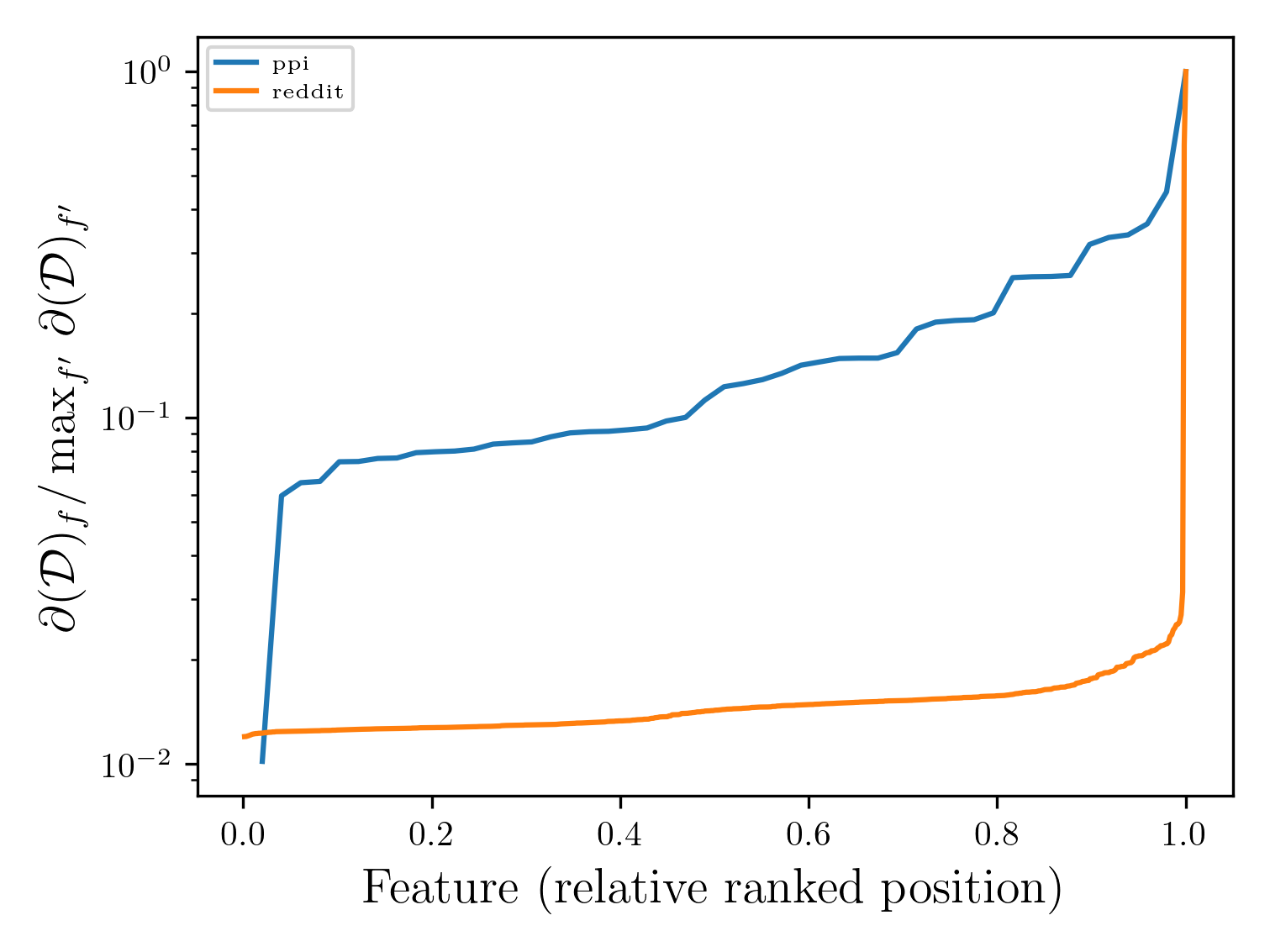}
    \caption{(Approximated) normalized intrinsic dimensionality.}
    \label{fig:graphsage_dimensions}
  \end{subfigure}
  \begin{subfigure}{.49\textwidth}
    \includegraphics[width=\linewidth]{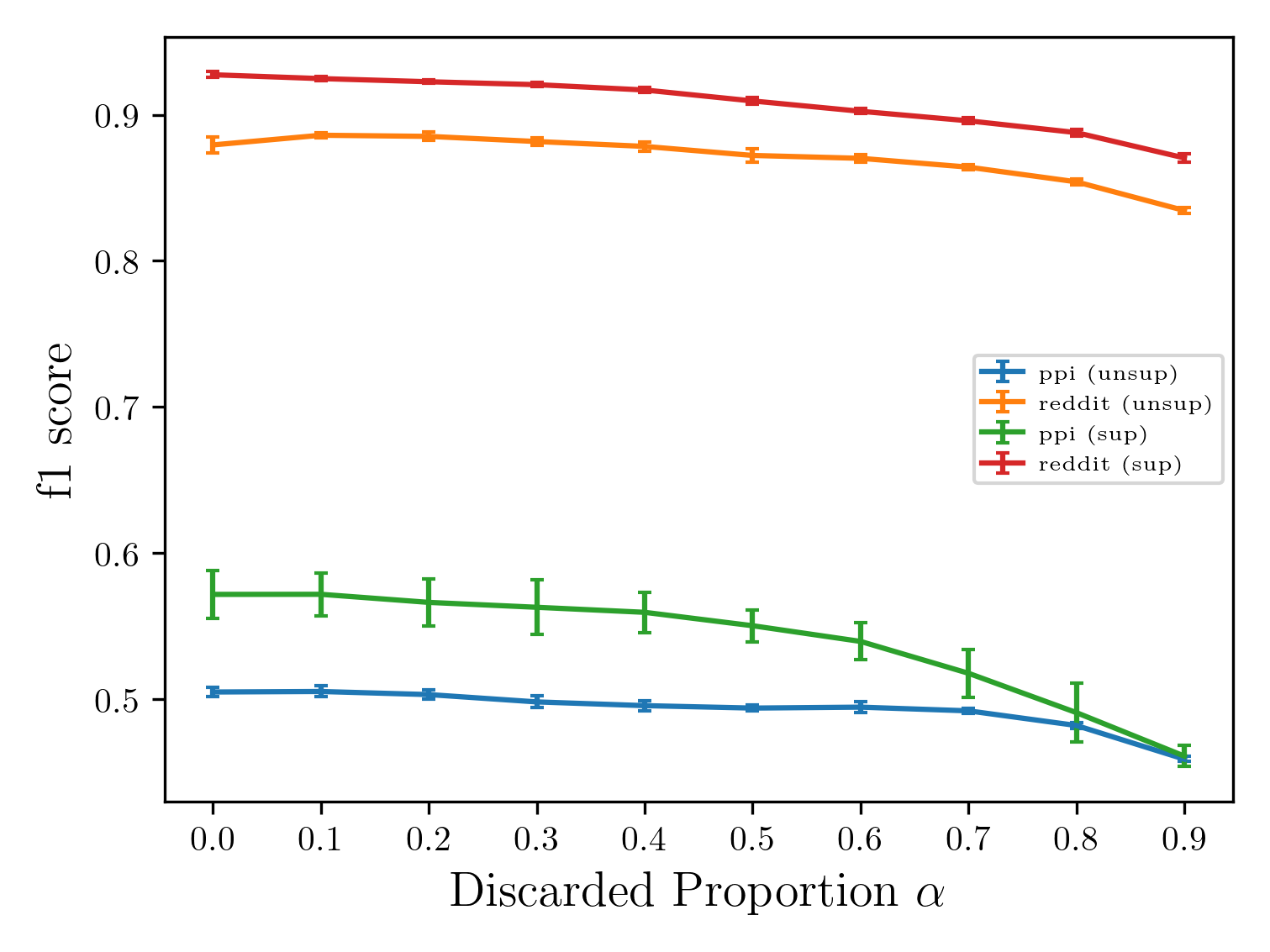}
    \caption{Performance of resulting model.}
    \label{fig:graphsage_factors}
  \end{subfigure}
  \caption{Experiment based on \GraphSage.}
  \label{fig:graphsage_experiment}
\end{figure}

\begin{figure}[H]
  \centering
  \begin{subfigure}{.49\textwidth}
    \includegraphics[width=\linewidth]{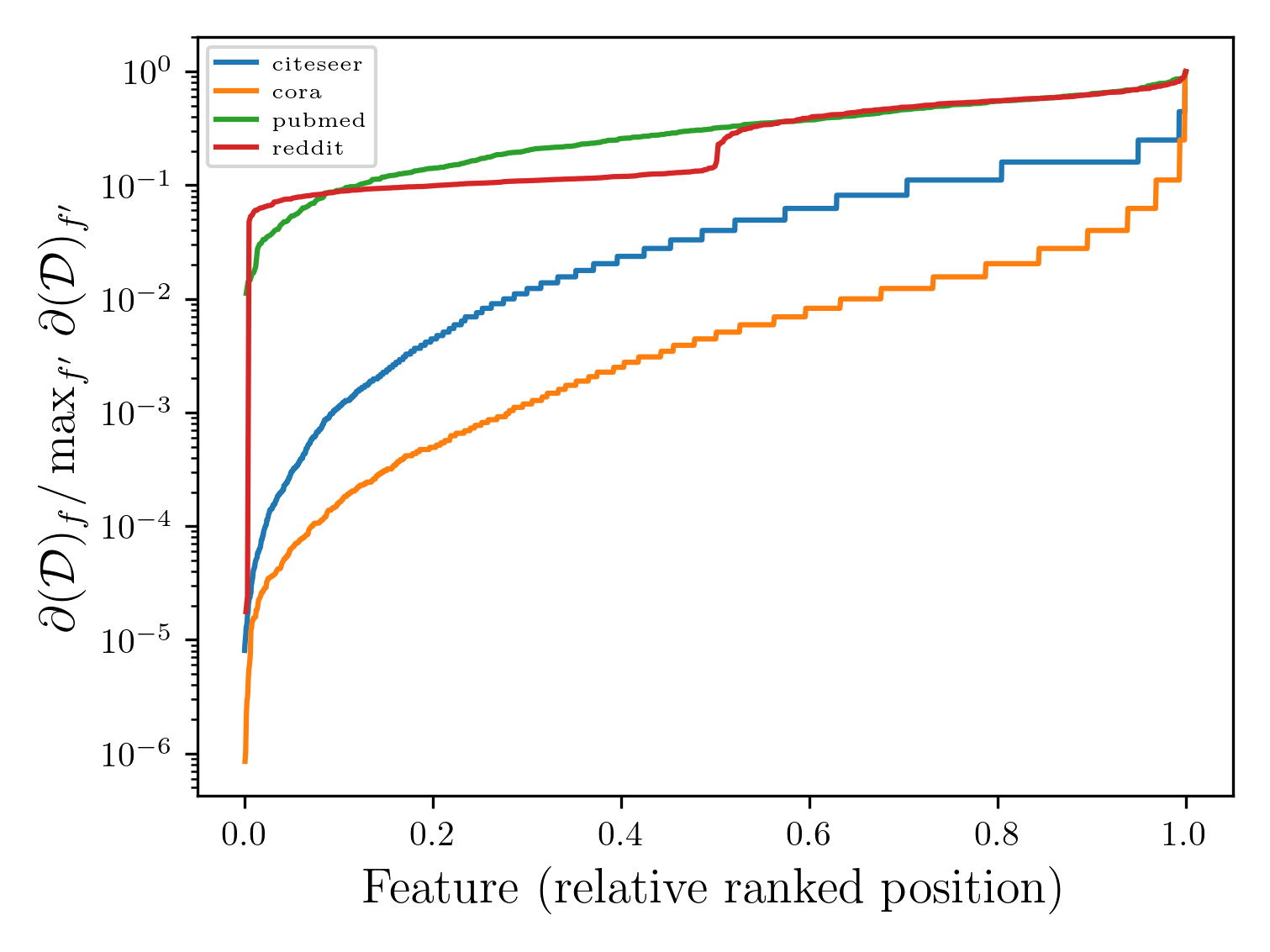}
    \caption{(Approximated) normalized intrinsic dimensionality.}
    \label{fig:sgc_dimensions}
  \end{subfigure}
  \begin{subfigure}{.49\textwidth}
    \includegraphics[width=\linewidth]{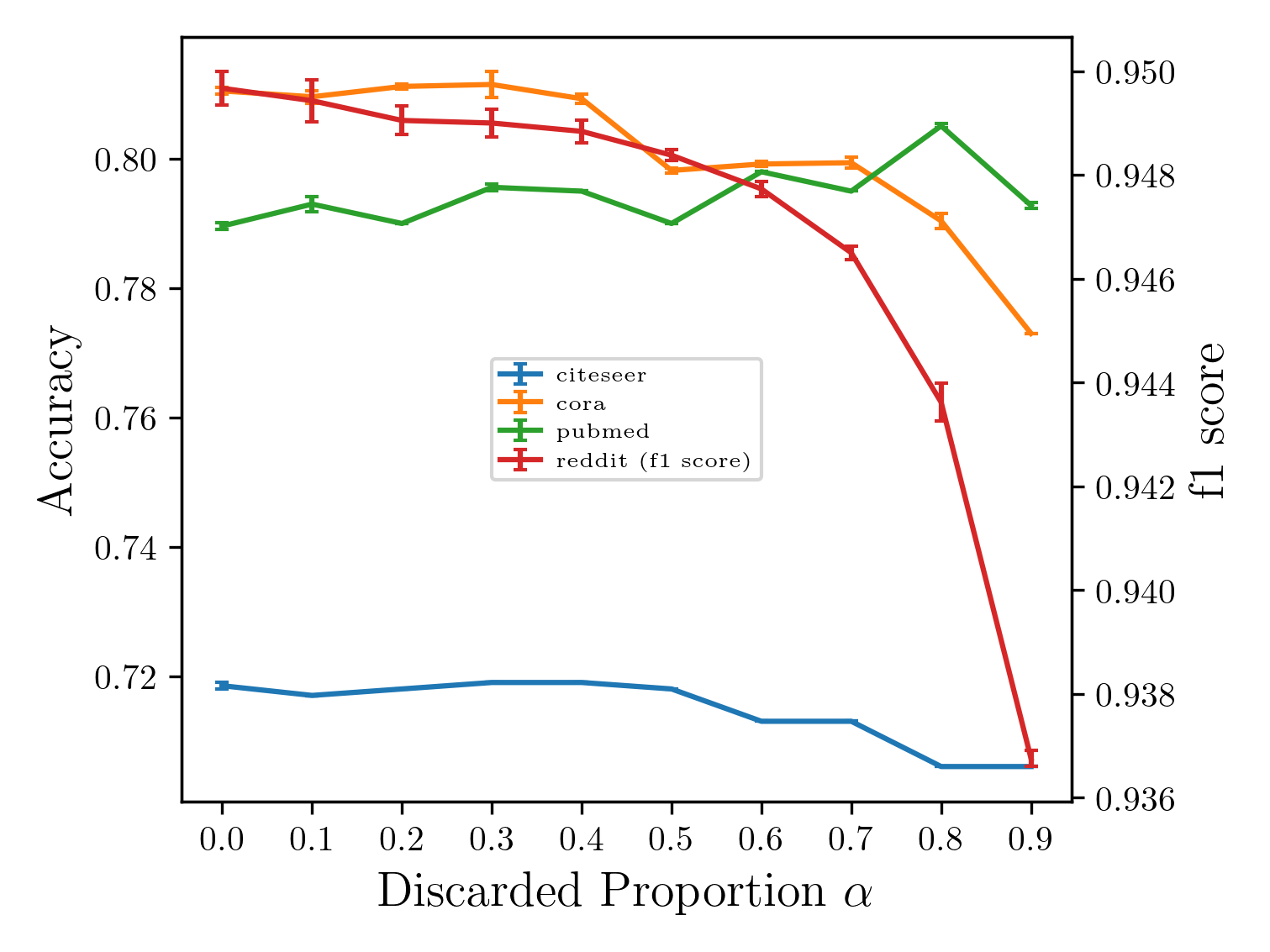}
    \caption{Performance of resulting model.}
    \label{fig:sgc_factors}
  \end{subfigure}
  \caption{Experiment based on \SGC.}
  \label{fig:sgc_experiment}
\end{figure}


\end{document}